\documentclass{article}

\usepackage{microtype}
\usepackage{graphicx}
\usepackage{subcaption}
\usepackage{booktabs} %

\usepackage{hyperref}

\usepackage[preprint]{icml2026}

\usepackage{amsmath}
\usepackage{amssymb}
\usepackage{mathtools}
\usepackage{amsthm}
\usepackage{booktabs}
\usepackage{multirow}
\usepackage{array}
\usepackage{colortbl}
\usepackage{xcolor}

\usepackage[capitalize,noabbrev]{cleveref}

\theoremstyle{plain}

\theoremstyle{definition}

\theoremstyle{remark}

\usepackage[textsize=tiny]{todonotes}

\icmltitlerunning{Language Generation as Optimal Control: Closed-Loop Diffusion in Latent Control Space}

\makeatletter
\renewcommand{\paragraph}{%
  \@startsection{paragraph}{4}%
  {\z@}{0.5em}{-0.5em}%
  {\normalfont\normalsize\bfseries}%
}

\begin{document}

\twocolumn[
  \icmltitle{Language Generation as Optimal Control: \\Closed-Loop Diffusion in Latent Control Space}

  \icmlsetsymbol{equal}{*}
  \begin{icmlauthorlist}
    \icmlauthor{ZiYi Dong}{equal,sysu}
    \icmlauthor{Yuliang Huang}{equal,sysu}
    \icmlauthor{Weijian Deng}{anu}
    \icmlauthor{Xiangyang Ji}{tsinghua}
    \icmlauthor{Liang Lin}{sysu}
    \icmlauthor{Pengxu Wei}{sysu}
  \end{icmlauthorlist}

  \icmlaffiliation{sysu}{Sun Yat-sen University, Guangzhou, Guangdong, China}
  \icmlaffiliation{anu}{Australian National University, Canberra, ACT, Australia}
  \icmlaffiliation{tsinghua}{Tsinghua University, Beijing, China}

  \icmlcorrespondingauthor{Pengxu Wei}{weipx3@mail.sysu.edu.cn}

  \icmlkeywords{Machine Learning, ICML}

  \vskip 0.3in
]

\printAffiliationsAndNotice{\icmlEqualContribution}

\begin{abstract}

This work reformulates language generation as a stochastic optimal control problem, providing a unified theoretical perspective to analyze autoregressive and diffusion models and explain their limitations (\textit{Efficiency-Fidelity Paradox, Irreversibility Error Propagation, Optimization Tractability and Fidelity}) in terms of combination of trajectory singularity, adjoint state vanishing, and gradient absence.
To address these issues, we approximate the solution to the Hamilton-Jacobi-Bellman (HJB) equation, yielding an optimal policy that acts as a closed-loop controller.
To bypass the intractability of directly solving the HJB PDE, we employ Flow Matching as the optimal trajectory solver within the rectified latent control space. This allows our Manta-LM with Global Integral Operator to approximate the global vector field, effectively realizing a model that simultaneously achieves high-fidelity text generation and efficient, low-cost parallel sampling.
Empirically, our method achieves strong performance on language modeling and conditional generation tasks, while exhibiting improved stability, efficiency, and controllability.
\end{abstract}

\section{Introduction}

The paradigm of Large Language Models (LLMs) has been dominated by Autoregressive Models (ARMs) in a sequential, ``next-token" prediction manner~\cite{brown2020language, touvron2023llama}, demonstrating remarkable scaling properties and emergent reasoning capabilities. The emerging Diffusion Language Models (DLMs) break serial constraints as a promising competitor to ARMs and promise a global receptive field and parallelizable sampling for generation. Discrete DLMs, \emph{e.g.}, D3PM~\cite{austin2021structured}, MDLM~\cite{sahoo2024simple}, and LLaDA~\cite{LLaDA}, define a forward corruption process via transition matrices, \emph{e.g.}, masking, within a discrete token space, and subsequently train a model by predicting the original tokens or the denoising trajectory. Continuous DLMs, \emph{e.g.}, CDCD~\cite{dieleman2022continuous}, and RDLM~\cite{RDLM}, project discrete tokens into a continuous embedding space to apply standard Gaussian diffusion frameworks, training the model to denoise high-dimensional vectors that are ultimately mapped back to discrete vocabulary.

Despite great progress, several crucial issues challenge those generative language paradigms, critically defining the current technological ceiling of LLMs. 
\emph{1) Efficiency-Fidelity Paradox:} 
ARMs are architecturally tethered to serial decoding, creating a linear computational bottleneck ($O(N)$). While DLMs theoretically allow for parallel refinement, their reliance on heuristic Gaussian denoising necessitates hundreds of resampling steps to converge, merely trading one efficiency bottleneck for another. 
\emph{2) Irreversibility Error Propagation:} ARMs in generation suffer from accumulated errors. In the open-loop setting of ARMs, a single "hallucinated" token in a sequence becomes the fixed prior for all subsequent steps, creating an accumulative and irreversible drift from the optimal trajectory. 
Standard DLMs similarly struggle with this situation: without a global guidance mechanism to regulate the denoising flow, the latent state often drifts into low-density semantic regions. %
\emph{3) Optimization Tractability and Fidelity:} Recent attempts to apply continuous diffusion directly in the token embedding space \cite{strudel2022self, dieleman2022continuous} face a geometric hurdle. The raw embedding space is sparse and clustered rather than being a smooth manifold. This ``ill-conditioned topology" results in high-curvature generation trajectories and severe quantization errors, where the continuous denoising process struggles to map back to discrete linguistic tokens without losing structural integrity.

To address those issues, we first reformulate generative language models and revisit those models from a stochastic optimal control perspective. Then, we propose \textbf{Manta-LM}, which connects text generation with continuous dynamical systems. Our approach rests on two theoretical pillars: \textit{Manifold Rectification} and \textit{Optimal Closed-Loop Control}. 
First, we employ a regularized Variational Autoencoder (VAE) to map the ill-conditioned discrete space to a compact, locally Euclidean latent manifold. This rectification reduces topological stiffness and encourages smoother transport trajectories.
Second, within this continuous latent space, we model generation as an \emph{Optimal Control} problem \cite{benamou2000computational}. By approximating the Hamilton-Jacobi-Bellman (HJB) equation via \emph{Flow Matching} \cite{lipman2022flow,Bertucci2023StochasticOT}, our model learns a vector field that acts as a \textit{Closed-Loop Feedback Controller}.

\section{Related Works}
\textbf{Autoregressive Language Models.} 
Autoregressive (AR) models have long been the dominant paradigm in language modeling, forming the backbone of modern large language models such as GPT \cite{brown2020language} and LLaMA \cite{touvron2023llama}. By factorizing the joint distribution of a sequence into a product of conditional probabilities, AR models excel at modeling local syntactic dependencies and achieve strong likelihood-based performance. 
However, the AR paradigm fundamentally enforces a strictly sequential generation process, where each token decision is irrevocable once sampled. This token-level hard commitment renders inference inherently non-parallelizable and amplifies error accumulation through exposure bias. Recent analyses have increasingly recognized these limitations, motivating alternatives that relax strict left-to-right decoding in favor of global or iterative refinement strategies. Our work departs from AR generation by reinterpreting this paradigm through the lens of control theory, identifying AR decoding as a form of greedy open-loop control operating in a discrete state space, which lacks mechanisms for global trajectory optimization or feedback correction.

\textbf{Discrete Diffusion Language Models.} 
Discrete DLMs extend diffusion-based generative modeling to categorical data by defining Markovian noising and denoising processes over discrete token spaces \cite{sohl2015deep, hoogeboom2021argmax, austin2021structured}. Among these, D3PM \cite{austin2021structured} establishes a general framework using arbitrary transition matrices, while subsequent works explore masked diffusion as a particularly effective instantiation for language \cite{sun2022score, lou2023discrete, shi2024simplified, sahoo2024simple, ou2024your}. 
Recent advances demonstrate that discrete DLMs can achieve competitive perplexity with autoregressive models at GPT-2 scale, especially when incorporating absorbing states, score entropy objectives (SEDD), or refined masking schedules \cite{lou2023discrete, ou2024your, nie2025large}. Large-scale efforts further scale masked diffusion to billions of parameters and extend it to multimodal generation \cite{gong2024scaling, ye2025dream, swerdlow2025unified, yang2025mmada, li2025lavida}. 
Despite these successes, discrete diffusion models inherit fundamental limitations from the non-metric nature of token spaces. The absence of a differentiable geometry prevents the definition of meaningful gradients over token trajectories, complicating the application of score matching and optimal transport principles. Thus, many methods rely on heuristic masking, remasking, or conditional independence assumptions, leading to trade-offs between generation quality, stability, and efficiency. Those issues account for the inability of discrete models to perceive or optimize smooth generation trajectories.

\textbf{Continuous Diffusion Language Models.} 
To recover differentiability, several works embed discrete tokens into continuous spaces and apply diffusion processes therein. Early approaches diffuse word embeddings directly and discretize outputs via nearest-neighbor or thresholding operations \cite{li2022diffusion, dieleman2022continuous, gong2023diffuseq, gong2023diffuseqv2}. While conceptually simple, such methods often suffer from information loss during dequantization and struggle to preserve categorical semantics. 
More structured continuous relaxations operate on probability simplices or logit spaces, leveraging Dirichlet priors, simplex geometry, or concrete distributions to impose statistical constraints on the diffusion process \cite{han2023ssd, mahabadi2024tess}. Flow-matching and score-based techniques further interpret the simplex as a statistical manifold, enabling continuous-time modeling \cite{cheng2024categorical}. Nevertheless, these approaches generally underperform discrete diffusion in generation fidelity or incur substantial computational overhead, particularly at scale \cite{gulrajani2023likelihood}.

\section{Revisiting Generative Language Models from Stochastic Optimal Control}
\label{sec:theory}

In this section, we first cast generative modeling as a \textit{Stochastic Optimal Control} problem~\cite{fleming2012deterministic}, and present an elaborated theoretical analysis to examine existing generative language models (\emph{i.e.}, ARMs and DLMs). 
With stochastic optimal control, we formalize the generation process as the time evolution of a probability density under a vector field and provide theoretical explanations for the limitations of existing methods.

\subsection{Stochastic Optimal Control}
\label{sec:soc_framework}

Stochastic Optimal Control (SOC) is a control theory that finds a control law to drive a system's evolution with minimum cost in the presence of random noise. 
In the context of language generation, the generative process as the controlled evolution of a state $\mathbf{z}_t$ on a manifold $\mathcal{M}$ over a finite time horizon $t \in [0, 1]$. The dynamics are governed by the stochastic differential equation~\cite{benamou2000computational}:
\begin{equation}
    d\mathbf{z}_t = \mathbf{u}(\mathbf{z}_t, t) dt + \sigma(t) d\mathbf{w}_t, \quad \mathbf{z}_0 \sim p_{\text{prior}},
    \label{eq:sde_dynamics}
\end{equation}
where $\mathbf{u}(\cdot)$ is the \textit{control law} (vector field) to be learned, and $\sigma(t)$ modulates the exploration noise.
The objective of the generator is to transport the prior $p_{\text{prior}}$ to the data distribution $p_{\text{data}}$ with minimal effort. Following the Benamou-Brenier formulation \cite{benamou2000computational}, the optimal control policy $\mathbf{u}^*$ is obtained by minimizing the transport cost functional $J(\mathbf{u})$:
\begin{equation}
    J(\mathbf{u}) = \underbrace{\mathbb{E}_{\mathbf{z}_1 \sim p_1} [-\log p_{\theta}(\mathbf{z}_1)]}_{\text{Terminal Cost (Data Fidelity)}} + \lambda \int_0^1 \underbrace{\mathbb{E}_{\mathbf{z}_t \sim p_t} \left[ \frac{1}{2} \|\mathbf{u}_t(\mathbf{z}_t)\|^2 \right]}_{\text{Running Cost (Kinetic Energy)}} dt.
    \label{eq:hjb}
\end{equation}

According to \textit{dynamic programming principle}, the optimal control law $\mathbf{u}^*$ can be characterized by the gradient of the \emph{optimal value function} $V(\mathbf{z}, t) = \inf_{\mathbf{u}} \mathbb{E} \left[ J(u) \right]$ (the minimum cost-to-go), which satisfies  \textit{Hamilton-Jacobi-Bellman (HJB)} equation. The optimal controller is necessarily:
\begin{equation}
    \mathbf{u}^*(\mathbf{z}, t) = -\nabla_{\mathbf{z}} V(\mathbf{z}, t).
    \label{eq:optimal_control_law}
\end{equation}
This indicates two critical properties for an ideal control system: \emph{1) Closed-Loop Feedback:} The control $\mathbf{u}^*$ depends on the \textit{current global state} $\mathbf{z}_t$, using the potential landscape $\nabla_{\mathbf{z}} V$ to correct deviations. \emph{2) Smooth Geodesic Flow:} Minimizing the kinetic energy term $\frac{1}{2}\|\mathbf{u}\|^2$ mandates that the optimal trajectory follows low-energy, geodesic-like trajectories in the Wasserstein space.

More broadly, this view makes text generation comparable through three control properties: whether future objectives can influence current updates, whether the model has corrective directions toward valid states, and whether the trajectory geometry is sufficiently smooth for stable integration. These properties are especially relevant for boundary-conditioned generation, where a model must satisfy global constraints rather than only extend a prefix.

\subsection{Generative Language Modeling via SOC}
\label{sec:baselines_math}
With the optimal controller defined in  \cref{eq:optimal_control_law}, we next place representative language generation paradigms into the same state-update view.

\paragraph{I. Autoregressive Models (AR): Impulsive Open-Loop Control.}
Autoregressive generation operates in a limit of zero viscosity ($\sigma(t) \to 0$), where the state evolution is driven by discrete token updates rather than a continuous flow. Under SOC, the generative dynamics are modeled as a deterministic system driven by an \textit{impulsive control law}:
\begin{equation}
    d\mathbf{z}_t = \underbrace{\left[ \sum_{k=1}^N \mathbf{f}_{\theta}(\mathbf{z}_{t_k}, \mathbf{h}_{t_k}) \cdot \delta(t - t_k) \right]}_{\text{Impulsive Control } \mathbf{u}_{\text{AR}}(\mathbf{z}_t, t)} dt + \underbrace{0}_{\text{No Diffusion}} \cdot d\mathbf{w}_t.
    \label{eq:ar_sde}
\end{equation}
where $\delta(\cdot)$ is Dirac delta function, $\mathbf{h}_{t_k}$ is the past hidden state and $\mathbf{f}_{\theta}$ is the greedy update at step $k$. Under this continuous-time interpretation, AR sampling corresponds to a \textit{singular trajectory}, a piecewise constant path with impulsive updates at transition points $t_k$. The control update $\mathbf{f}_{\theta}$ is derived from local likelihood maximization, lacking adjoint state feedback for global optimality. Thus, under the SOC abstraction, ARMs behave as a stiff, open-loop control system that is prone to error accumulation when global terminal constraints matter.

\paragraph{II. Discrete DLMs: Gradient-Free Rate Control.}
Discrete DLMs operate on a categorical lattice where the state evolution follows a Controlled Continuous-Time Markov Chain (CTMC). In our SOC formulation, this is governed by a \textit{Jump SDE} driven by Poisson counter processes $N$:
\begin{equation}
    d\mathbf{z}_t = \sum_{\mathbf{y} \in \mathcal{D}, \mathbf{y} \neq \mathbf{z}_t} (\mathbf{y} - \mathbf{z}_t) \cdot dN_t\left( \lambda = [\mathbf{u}_t{\text{}}(\mathbf{z}_t)]_{\mathbf{z}_t \to \mathbf{y}} \right).
    \label{eq:disc_sde}
\end{equation}
Here, the control law $\mathbf{u}_{\text{}}$ does not define a velocity vector, but rather a \textit{Transition Rate Tensor} that modulates the intensity $\lambda$ of jumping to a neighbor state $\mathbf{y}$. 
Crucially, the state difference $(\mathbf{y} - \mathbf{z}_t)$ represents a discrete hop on a Hamming graph, not a tangent vector in a Riemannian manifold. 
Thus, the smooth differential operator $\nabla_{\mathbf{z}}$ required by the continuous HJB formulation is not directly available. Discrete methods may still define useful graph-based or finite-difference surrogate scores, but these are not the same object as a smooth vector field on a continuous manifold. Under the SOC lens, the system therefore relies on stochastic transition search rather than geometric guidance from a differentiable descent direction.

\paragraph{III. Continuous DLMs: Ill-Conditioned Control.}
Existing continuous DLMs that apply diffusion directly to raw word embeddings restore the functional form of the optimal control SDE but can be limited due to topological constraints. The dynamics follow the standard reverse-time SDE: 
\begin{equation}
    d\mathbf{z}_t = \underbrace{\left[ \mathbf{f}(\mathbf{z}, t) - g(t)^2 \nabla_{\mathbf{z}} \log p_t(\mathbf{z}) \right]}_{\text{Feedback Control } \mathbf{u}_{\text{}}(\mathbf{z}, t)} dt + g(t) d\mathbf{w}_t.
\end{equation}
Although $\mathbf{u}_{\text{}}$ takes the form of a closed-loop controller via the score function, the state space $\mathcal{Z}_{\text{emb}}$ is sparse, clustered, and non-manifold. This irregular geometry can induce large variations in the score function, resulting in a vector field with high local Lipschitz constants. In control theory, this characterizes a \textit{stiff system}, which forces numerical solvers to take smaller steps to maintain stability and can reduce the efficiency benefits of continuous modeling.

\subsection{Diagnosing Generative Dynamics in Autoregressive and Diffusion Flows}
\label{sec:pathologies}
By contrasting the generative dynamics (\cref{eq:ar_sde}, \ref{eq:disc_sde}) against the optimal control law $\mathbf{u}^* = -\nabla V$ (\cref{eq:optimal_control_law}), we diagnose three issues under the SOC lens: \textit{Trajectory Singularity}, \textit{Adjoint State Vanishing}, and \textit{Gradient Absence}. These issues help explain why current paradigms can face serial inefficiency, irreversible error propagation, and optimization intractability in globally constrained generation settings.

\noindent
\textbf{I. Efficiency Paradox via Trajectory Singularity and System Stiffness.}
Optimal control favors finite kinetic energy $\mathcal{A} = \frac{1}{2} \int_0^1 \|\mathbf{u}_t\|^2 dt < \infty$. However, substituting the control laws derived in \cref{sec:baselines_math} reveals how several language-modeling paradigms depart from this smoothness condition under the continuous-control interpretation.

\emph{i) Impulsive dynamics of AR:} 
Substituting the impulsive control law from \cref{eq:ar_sde} yields a divergence:
    \begin{equation}
        \mathcal{A}_{\text{AR}} \propto \int_0^1 \left\| \sum \mathbf{f}_k \delta(t-t_k) \right\|^2 dt \to \infty.
    \end{equation}
The $L_2$ norm of the Dirac delta is not finite, so this continuous-time embedding corresponds to a highly stiff path with impulsive curvature. Such dynamics are poorly matched to parallel continuous solvers (\textit{e.g.}, Picard iteration), reflecting the serialism bottleneck of AR decoding.
    
\emph{ii) Lipschitz Explosion of Continuous DLM:} 
    For diffusion on raw embeddings, the sparse, non-manifold topology can induce large variations in the score function. The local Lipschitz constant $L$ explodes:
    \begin{equation}
        L = \sup_{\mathbf{z} \neq \mathbf{y}} \frac{\| \mathbf{u}(\mathbf{z}) - \mathbf{u}(\mathbf{y}) \|}{\| \mathbf{z} - \mathbf{y} \|} \gg 1.
    \end{equation}
    Since numerical stability requires step sizes $\Delta t < 2/L$, a large $L$ forces solvers to take smaller steps ($\Delta t \to 0$). This increases the Number of Function Evaluations (NFE), reducing the practical efficiency of continuous modeling and making optimization less tractable.

\textbf{II. Irreversibility Error Propagation via Adjoint State Vanishing (Open-Loop).}
Optimal policies depend on the \textit{Prediction Horizon} $H$, which defines the scope of the objective functional $J_H(\mathbf{u}_t) = \mathbb{E}[\int_t^{t+H} \frac{1}{2}\|\mathbf{u}\|^2 d\tau + \Phi(\mathbf{z}_{t+H})]$. The global guidance is carried by the \textit{Adjoint State} $\mathbf{p}_t$, which back-propagates the terminal potential $\Phi$ from $t+H$.
    
    \emph{i) Adjoint Vanishing:} Global optimality requires a full horizon $H = 1-t$. In contrast, AR effectively operates like an MPC policy with a \emph{single-step horizon} ($H \to 0$ in the continuous limit). This truncation discards the integral cost beyond the immediate step, severing the backward link to $\Phi(\mathbf{z}_1)$ and forcing $\mathbf{p}_t \equiv \mathbf{0}$.
    
    \emph{ii) Lyapunov Instability:}
    Without the restoring force provided by the adjoint feedback $\mathbf{u}_{\text{fb}} \propto \mathbf{p}_t$, the system operates in an unstable open-loop regime. Small quantization errors $\epsilon$ amplify over time rather than being suppressed, leading to diverging error dynamics.%

This provides a control-theoretic explanation for \textbf{Irreversible Error Propagation} (\textit{e.g.}, hallucinations) in AR, where early mistakes can steer subsequent generation away from the desired trajectory.

\textbf{III. Geometric Blindness via Gradient Absence.}
Efficient optimal control relies on a \textit{Gradient Flow} $\mathbf{u}^* = -\nabla_z V$ to guide the state towards high-density regions. However, discrete diffusion models (\cref{eq:disc_sde}) are constrained to a lattice~$\mathcal{D}$ equipped with the Hamming metric.
    \emph{i) Metric Singularity:} 
    The discrete space $\mathcal{D}$ admits no tangent space $T_{\mathbf{z}}\mathcal{M}$. Consequently, the differential operator $\nabla_{\mathbf{z}}$ required by the HJB equation is ill-defined. 
    
    \emph{ii) Fidelity Degradation:}
    Without the directional guidance of a smooth gradient field $-\nabla_z V$, the controller relies on stochastic transition search. Compared with continuous flows that can follow a differentiable descent direction toward the data manifold, this search can settle for sub-optimal local states, helping explain the \emph{fidelity gap} and lack of fine-grained control in discrete generative models (\textit{e.g.}, AR and DLM).

This geometric limitation contributes to the \textbf{Efficiency-Fidelity Paradox}: improving sample quality often requires more stochastic refinement steps, which reduces the intended parallel speedup.

\begin{figure}[t]
    \centering
    \includegraphics[width=0.48\textwidth]{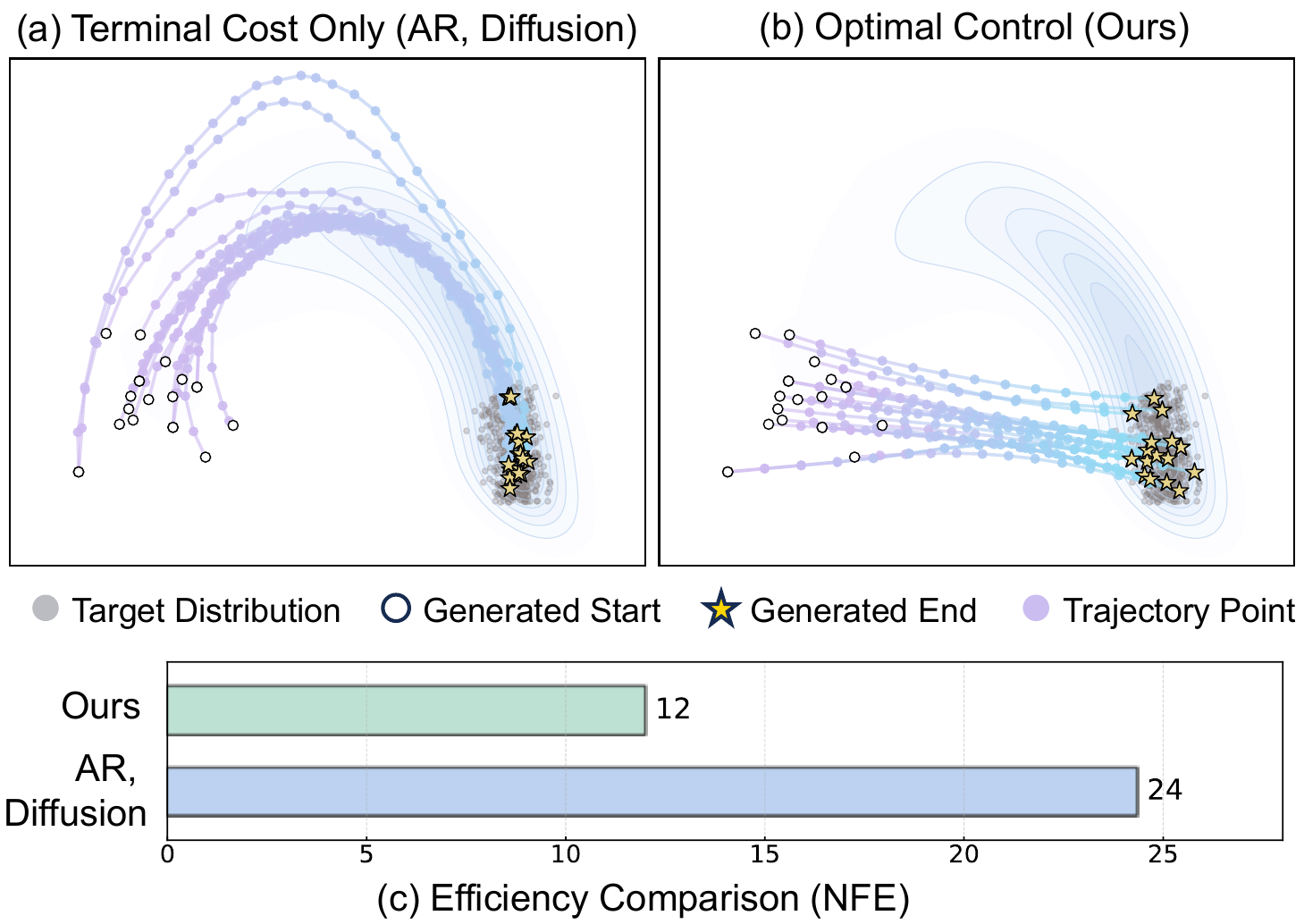}
    
    \caption{\textbf{Generation dynamics.} On a non-convex manifold, (a) AR and Diffusion are trapped in a slow, myopic crawl along the high-curvature density ridge. (b) In contrast, our method approximates the global optimal trajectory, bypassing curvature via the rectified latent geometry (energy-minimizing geodesic) for improved efficiency.}
    \vspace{-8pt}
    \label{fig:myopic}
\end{figure}

\begin{figure}[t]
    \centering
    \includegraphics[width=0.48\textwidth]{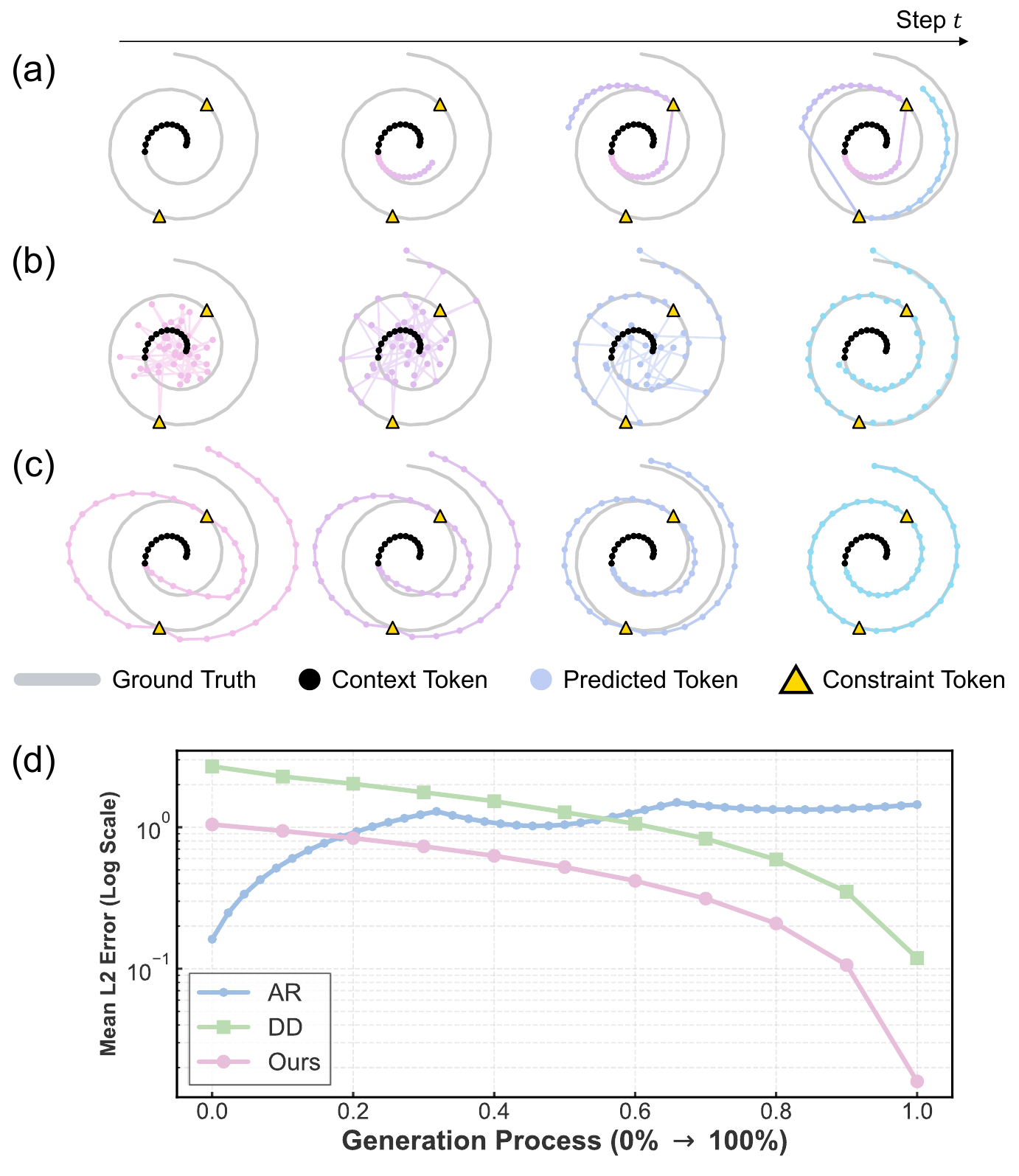}
    
    \caption{
    \textbf{Visualizing Generative Dynamics and Error Propagation on BVP task.} 
    Color from pink to blue denotes generation progress. 
    \textbf{(a)} AR suffers from \textit{compounding errors} (blue lines in (d)) due to open-loop myopia, drifting off-manifold. 
    \textbf{(b)} Discrete DLM relies on stochastic combinatorial search showing jagged trajectories caused by \textit{geometric blindness} (lack of gradients). 
    \textbf{(c)} Our Manta-LM acts as an optimal closed-loop controller, utilizing the learned vector field to self-correct deviations along a low-energy geodesic.
    \textbf{(d)} Quantitative analysis shows that Manta-LM achieves superior convergence stability and reduces terminal error.
}
    \vspace{-10pt}
    \label{fig:acc_err}
\end{figure}

\section{Optimal Controller as Diffusion Language Model}
\label{sec:latent_flow_hjb}

The SOC analysis above suggests three requirements for an effective closed-loop language generator: a smooth control space in which continuous dynamics are tractable, locality-preserving controllability so partial constraints can be imposed and corrected locally, and global interaction so the controller can use full-sequence information when updating each state. To instantiate these requirements, we propose \textbf{Manta-LM}, a Latent Diffusion Language Model formulated as an optimal closed-loop controller.
Manta-LM uses a regularized TextVAE to construct the latent control space, a locality-preserving convolutional encoder to maintain local controllability, and a Transformer controller to model global interactions. Within this latent space, Flow Matching provides a practical trajectory-level approximation to the optimal controller in~\cref{eq:optimal_control_law}, targeting high data fidelity with low inference cost in~\cref{eq:hjb}.

\subsection{Control-Friendly Manifold Rectification}

\textbf{Rectification via Diffeomorphism.}
We introduce a Variational Autoencoder (VAE) as a coordinate transformation $\psi: \mathcal{D} \to \mathcal{Z}$, regularized by a Gaussian prior via $\mathcal{L}_{\text{KL}}$. This regularization encourages $\mathcal{Z}$ to be \textbf{diffeomorphic to Euclidean space}, effectively performing \textit{Manifold Rectification} on the non-metric token space. By constructing a continuous, locally Euclidean latent representation, we obtain a setting where the gradient operator $\nabla_{\mathbf{z}}$ is well-defined and the learned dynamics can be more Lipschitz regular. This yields a \textbf{Control-Friendly Geometry}, satisfying the theoretical prerequisites for stable, unique ODE~\cite{Bullo2004GeometricCO} solutions within our Optimal Control framework.

\textbf{Topology-Preserving Compression.}
Transformer-based VAEs suffer from global information entanglement, where each latent vector aggregates information from the entire sequence. This destroys the spatial locality required for controllable generation, rendering masked in-painting mathematically ill-posed due to inevitable information leakage. We resolve this by employing Local Integral Operators to enforce strict spatial spatial disentanglement.

\begin{figure}[t]
    \centering
    \includegraphics[width=0.48\textwidth]{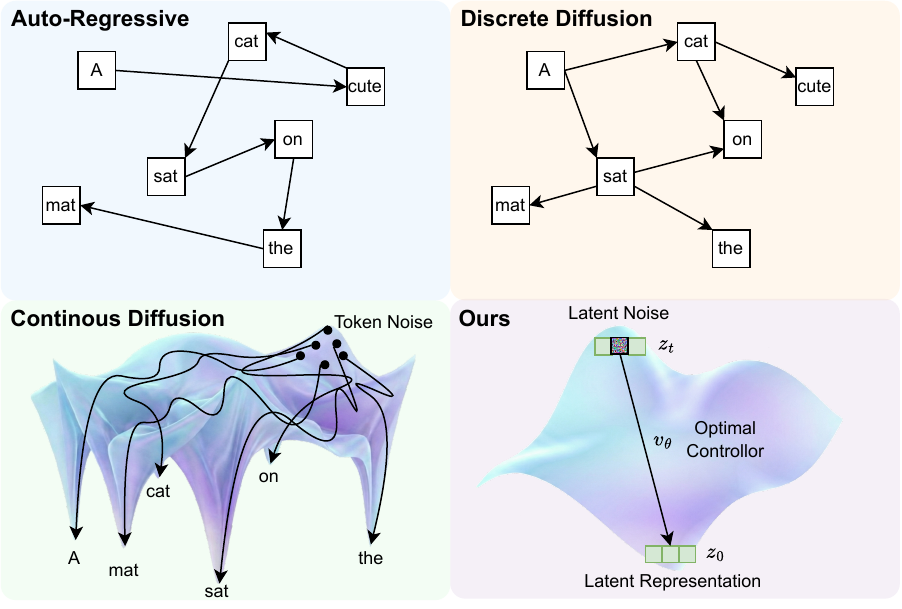}
    
    \caption{\textbf{Geometric comparison.} Unlike (a) Autoregressive models' serial paths or (b-c) Diffusion baselines' high-curvature trajectories in ill-conditioned spaces, (d) Ours operates on a rectified latent manifold. The learned optimal vector field $v_{\theta}$ enables energy-minimizing, straight-line transport from noise to data.}
    \vspace{-8pt}
    \label{fig:diffusion}
\end{figure}

\subsection{Flow Matching as the Lagrangian Solver}
Directly solving the PDE in \cref{sec:soc_framework} is intractable. However, we leverage the duality between the Eulerian (PDE) and Lagrangian (Path) specifications.
Conditional Flow Matching (CFM) with Optimal Transport paths parameterizes the target trajectory as $\mathbf{z}_t = (1-t)\mathbf{z}_0 + t\mathbf{z}_1$. The velocity field of this path is constant: $\mathbf{u}_t(\mathbf{z} | \mathbf{z}_1) = \frac{d\mathbf{z}_t}{dt} = \mathbf{z}_1 - \mathbf{z}_0$.

\textbf{Approximate solution of HJB:} By the Benamou-Brenier theory \cite{benamou2000computational}, straight-line interpolation minimizes the kinetic energy action $\mathcal{A} = \int \|\mathbf{u}\|^2 dt$ in Wasserstein space.
Therefore, the marginal vector field learned by the network, $v_\theta^*(\mathbf{z}, t) = \mathbb{E}[\mathbf{u}_t | \mathbf{z}_t = \mathbf{z}]$, provides a tractable approximation to the HJB-induced dynamics~\cite{Bertucci2023StochasticOT}. Thus, minimizing the simple regression loss:
\begin{equation}
    \mathcal{L}_{\text{CFM}}(\theta) = \mathbb{E}_{t, \mathbf{z}_0, \mathbf{z}_1} \left\| v_\theta(\mathbf{z}_t, t) - (\mathbf{z}_1 - \mathbf{z}_0) \right\|^2,
\end{equation}
serves as a practical Lagrangian surrogate for the stochastic optimal control problem in rectified latent space. \cref{fig:acc_err}(c) indicates that this solution can be more efficient and accurate than AR and Discrete Diffusion.

\subsection{Transformer as the Global Integral Operator}
Finally, we analyze the architectural realization of the control law $\mathbf{u}^*$. Since $V(\mathbf{z})$ depends on the global configuration of the latent particles (tokens), the gradient $\nabla_{\mathbf{z}} V$ is a non-local operator.
We model the latent state as an interacting particle system. The total force on particle $i$ is given by the integral of pairwise interactions:
\begin{equation}
    \mathbf{u}^{(i)}(\mathbf{z}) = -\nabla_{\mathbf{z}^{(i)}} V \approx \int_{\Omega} \mathcal{K}(\mathbf{z}^{(i)}, \mathbf{z}^{(j)}) \mathbf{g}(\mathbf{z}^{(j)}) d\mathbf{z}^{(j)}.
\end{equation}
The Transformer Self-Attention mechanism is precisely the discrete approximation of this integral operator:
\begin{equation}
    \text{Attention}(\mathbf{z})_i = \sum_{j=1}^N \underbrace{\frac{\exp(\mathbf{W}_q \mathbf{z}^{(i)} (\mathbf{W}_k \mathbf{z}^{(j)})^{\top})}{\mathcal{Z}}}_{\text{Kernel } \mathcal{K}_{ij}} \underbrace{\mathbf{W}_v \mathbf{z}^{(j)}}_{\text{Force } \mathbf{g}_j}.
\end{equation}
Therefore, the Transformer is not an arbitrary choice; it is the structural discretization of the global gradient flow required by the HJB dynamics.

\begin{table}[t]
    \centering
    \caption{Theoretic comparison of generative dynamics.}
    \label{tab:theory_summary}
    \scriptsize
    \setlength{\tabcolsep}{1.5pt}
    \resizebox{\linewidth}{!}{
    \begin{tabular}{lcccc}
        \toprule
        \textbf{Property} & \textbf{AR} & \textbf{Discrete DLM} & \textbf{Continuous DLM} & \textbf{Manta-LM} \\
        \midrule
        Control Type & Open-loop & Rate control & Score feedback & \textbf{Closed-loop} \\
        Trajectory & Singular & Jump process & Stiff flow & \textbf{Smooth geodesic} \\
        Look-ahead & Myopic & Global/no smooth grad. & Global/ill-cond. & \textbf{Global/grad.} \\
        Energy Efficiency & Low & Medium & Low & \textbf{High} \\
        \bottomrule
    \end{tabular}
    }
\end{table}

\section{Experiments}

\subsection{Experimental Setup}

\noindent
\textbf{Zero-shot Language Modeling.}
To rigorously evaluate density estimation, the primary benchmark where Discrete DLMs seek to challenge AR models, we strictly adhere to the protocols established by recent discrete diffusion studies~\cite{lou2023discrete,ou2024your}. Specifically, models are trained on OpenWebText~\cite{Gokaslan2019OpenWeb} following SEDD's processing~\cite{lou2023discrete} and evaluated on \textbf{five} benchmarks: LAMBADA~\cite{paperno2016lambada}, WikiText-2/103~\cite{merity2016pointer}, PTB, and 1BW~\cite{radford2019language}. %

\noindent
\textbf{Conditional Text Generation.}
Following the evaluation pipeline of DiffuSeq~\cite{gong2023diffuseq}, models are evaluated on four representative conditional text generation tasks: 
Paraphrase Generation on QQP~\cite{quora-question-pairs};
\emph{Question Generation} on {Quasar-T}~\cite{dhingra2017quasar};
\emph{Text Simplification} on Wiki-Auto~\cite{jiang2020neural};
and \emph{Open-domain Dialogue} on CCD~\cite{zhou2018commonsense}.

\noindent
\textbf{Evaluation Metrics.}
Our evaluation focuses on both quality and diversity. Quantitatively, we measure generation quality via BLEU, ROUGE-L (R-L), and BERTScore (Score), while diversity is assessed using Distinct-1 (D-1). For zero-shot density estimation, we report Perplexity (PPL). For unconditional free generation, we additionally report generation perplexity (Gen PPL) together with unigram entropy.

\subsection{Main Results}

\paragraph{Zero-shot Language Modeling.}
As shown in \cref{tab:zero_shot}, our model consistently outperforms baselines, establishing a new state-of-the-art over discrete diffusion and AR models. On the massive 1BW dataset, we achieve a perplexity of 62.55, improving over the strong diffusion baseline MD4 (68.10) and the autoregressive GPT-2 baseline (75.2). Substantial reductions on LAMBADA (28\%) and WikiText-2 (12\%) further suggest that our latent optimal control architecture captures complex dependencies superior to prior discrete formulations, effectively bridging the gap to autoregressive density estimation.

\begin{table}[t]
\centering
\caption{ELBO-based zero-shot perplexity ($\downarrow$). Diffusion-based likelihoods are evaluated with the upper-bound/proxy protocol used in prior diffusion-LM work. The Manta-LM result uses the 101M model.}
\scriptsize
\setlength{\tabcolsep}{0pt}
\begin{tabular*}{\linewidth}{@{\extracolsep{\fill}}lccccc@{}}
\toprule
Method & Lambada & Wikitext2 & PTB & Wikitext103 & 1BW \\
\midrule
GPT-2 & \underline{45.04} & 42.43 & 138.43 & 41.6  & 75.2 \\
\midrule
D3PM & 93.47 & 77.28 & 200.82 & 75.16 & 138.92 \\
PLAID & 57.28 & 51.8  & 142.6  & 50.86 & 91.12 \\
SEDD & 50.92 & 41.84 & 114.24 & 40.62 & 79.29 \\
MD4 & 48.43 & \underline{34.94} & \underline{102.26} & \underline{35.90} & \underline{68.10} \\
RADD & 51.7  & 39.98 & 107.85 & 37.98 & 72.99 \\
AO-GPT & 54.92 & 42.24 & 123.49 & 41.51 & 76.73 \\
\midrule
Manta-LM & \textbf{34.80} & \textbf{30.58} & \textbf{97.76} & \textbf{35.35} & \textbf{62.55} \\
\bottomrule
\end{tabular*}
\label{tab:zero_shot}
\end{table}

For \cref{tab:zero_shot}, the reported diffusion-model PPL is an ELBO-based density-estimation proxy rather than direct free-generation likelihood. For Manta-LM, the bound decomposes into an encoder posterior term, a flow-prior density term computed by change of variables along the learned ODE, and a decoder likelihood term:
\begin{equation}
\begin{aligned}
\log p_\theta(x) \geq
\mathbb{E}_{q_\phi(\mathbf{z}\mid x)}
\big[&\log p_\theta(x\mid \mathbf{z}) + \log p_\theta(\mathbf{z}) \\
&- \log q_\phi(\mathbf{z}\mid x)\big].
\end{aligned}
\end{equation}
We compute $\mathrm{PPL}=\exp(-\mathcal{L}_{\mathrm{ELBO}}/N)$ over $N$ tokens, following the same zero-shot evaluation protocol as discrete diffusion baselines.

\begin{table}[t]
\centering
\caption{Unconditional generation quality and diversity. Gen PPL ($\downarrow$) measures free-generation fidelity, while unigram entropy ($\uparrow$) monitors diversity.}
\scriptsize
\setlength{\tabcolsep}{1.5pt}
\resizebox{\linewidth}{!}{
\begin{tabular}{llccccc}
\toprule
\textbf{Method} & \textbf{Metric} & \textbf{16} & \textbf{32} & \textbf{64} & \textbf{128} & \textbf{256} \\
\midrule
MDLM & Gen PPL & 137.02 & 120.14 & 103.37 & 82.64 & 66.28 \\
RADD & Gen PPL & 78.29 & 60.49 & 46.23 & 37.53 & 32.68 \\
\textbf{Manta-LM} & Gen PPL & \textbf{46.92} & \textbf{32.10} & \textbf{29.18} & \textbf{23.80} & \textbf{23.56} \\
\midrule
RADD & Entropy & 6.1076 & 6.3109 & 6.3501 & 6.4133 & 6.4214 \\
\textbf{Manta-LM} & Entropy & 6.1232 & 6.0459 & 5.7981 & 6.0382 & 5.9714 \\
\bottomrule
\end{tabular}
}
\label{tab:gen_ppl_entropy}
\end{table}

\begin{table*}[t]
\centering
\setlength{\tabcolsep}{3pt}
\caption{
\textbf{Method comparison.}
$^{\bullet}$ indicates AR models,
$^{\star}$ indicates Non-autoregressive Models and
$^{\ddagger}$ indicates Non-autoregressive Diffusion Models.
\textbf{Boldfaced} results show the best across all models;
\underline{Underlined} results show the best across all non-AR models.
}

\resizebox{\textwidth}{!}{
\begin{tabular}{l|c|cccc|cccc|cccc|cccc}
\toprule
\multirow{2}{*}{\textbf{Methods}} &
\multirow{2}{*}{\textbf{\# Params}} &
\multicolumn{4}{c|}{\textbf{Paraphrase}} &
\multicolumn{4}{c|}{\textbf{Question Generation}} &
\multicolumn{4}{c|}{\textbf{Text Simplification}} &
\multicolumn{4}{c}{\textbf{Open Domain Dialogue}} \\
\cmidrule(lr){3-6}
\cmidrule(lr){7-10}
\cmidrule(lr){11-14}
\cmidrule(lr){15-18}
& & BLEU & R-L & Score & D-1
  & BLEU & R-L & Score & D-1
  & BLEU & R-L & Score & D-1
  & BLEU & R-L & Score & D-1 \\
\midrule
GPT2-base FT$^{\bullet}$ & 117 M
& 0.198 & 0.521 & 0.825 & 0.980
& 0.074 & 0.271 & 0.605 & 0.960
& 0.308 & 0.546 & 0.802 & 0.944
& 0.011 & \textbf{0.151} & 0.528 & 0.919 \\

GPT2-large FT$^{\bullet}$ & 774 M
& 0.206 & 0.542 & 0.836 & \textbf{0.982}
& 0.111 & 0.322 & \textbf{0.635} & \textbf{0.967}
& 0.269 & 0.511 & 0.788 & 0.946
& 0.013 & 0.100 & \textbf{0.529} & 0.924 \\

GPVAE-T5$^{\bullet}$ & 220 M
& 0.241 & 0.589 & 0.847 & 0.969
& 0.125 & 0.339 & 0.631 & 0.938
& 0.339 & 0.583 & 0.817 & 0.931
& 0.011 & 0.101 & 0.432 & 0.563 \\

\midrule
LevT$^{\star}$ & 80 M
& 0.227 & 0.580 & 0.834 & 0.979
& 0.093 & 0.289 & 0.549 & 0.891
& 0.205 & 0.440 & 0.725 & \textbf{\underline{0.972}}
& 0.016 & 0.055 & 0.476 & \textbf{\underline{0.973}} \\

DiffuSEQ$^{\ddagger}$ & 91 M
& 0.241 & 0.588 & 0.837 & \underline{0.981}
& 0.173 & 0.367 & 0.612 & 0.906
& 0.362 & 0.585 & 0.813 & 0.926
& 0.014 & 0.106 & \underline{0.513} & 0.947 \\

SeqDiffuSeq$^{\ddagger}$ & 108 M
& 0.243 & - & 0.840 & \underline{0.981}
& 0.175 & - & \underline{0.617} & \underline{0.925}
& 0.371 & - & 0.821 & 0.908
& 0.011 & - & 0.443 & 0.961 \\

RDLM$^{\ddagger}$~\cite{RDLM} & -
& 0.215 & 0.482 & 0.704 & -
& 0.179 & 0.315 & 0.576 & -
& 0.280 & 0.467 & 0.687 & -
& 0.011 & 0.109 & 0.481 & - \\

\midrule
\rowcolor{gray!8}
\textsc{Manta-LM} & 101 M
& \cellcolor{gray!15}\textbf{\underline{0.301}}
& \cellcolor{gray!15}\textbf{\underline{0.653}}
& \cellcolor{gray!15}\textbf{\underline{0.848}}
& 0.971
& \cellcolor{gray!15}\textbf{\underline{0.186}}
& \cellcolor{gray!15}\textbf{\underline{0.378}}
& 0.613
& 0.880
& \cellcolor{gray!15}\textbf{\underline{0.389}}
& \cellcolor{gray!15}\textbf{\underline{0.613}}
& \cellcolor{gray!15}\textbf{\underline{0.822}}
& 0.904
& \cellcolor{gray!15}\textbf{\underline{0.020}}
& \cellcolor{gray!15}\underline{0.112}
& 0.485
& 0.933 \\
\bottomrule
\end{tabular}
}
\label{tab:cond}
\end{table*}

\paragraph{Conditional Text Generation.}
\cref{tab:cond} summarizes our performance against AR models and Non-AR models baselines. 
\emph{1) Quality \& AR-Parity:} \textsc{Manta-LM} sets a new non-autoregressive SOTA, surpassing SeqDiffuSeq by over 5 BLEU on Paraphrase Generation. Crucially, we achieve parity with or even surpass fine-tuned GPT2-large on Text Simplification and Question Generation, demonstrating ARM-level fidelity with Non-AR inference benefits.
\textbf{Diversity:} Our model maintains high Distinct-1 scores comparable to DiffuSeq. In high-entropy Open Domain Dialogue, we achieve the highest BLEU (1.95) among the compared methods, suggesting robustness in modeling complex one-to-many mappings.

\begin{table}[t]
\centering
\caption{
\textbf{Reconstruction and robustness of TextVAE.}
TextVAE achieves a bidirectional mapping between the token space and the latent control space. Reconstruction results demonstrate that it performs this mapping with high precision, and lower Lipschitz constants indicate strong robustness in the latent control space, allowing for a certain margin of error in the controller.
}
\footnotesize
\setlength{\tabcolsep}{0pt} 
\begin{tabular*}{\linewidth}{@{\extracolsep{\fill}}lcccc@{}} 
\toprule
& \multicolumn{2}{c}{\textbf{Reconstruction}} & \multicolumn{2}{c}{\textbf{Robustness}} \\
& \multicolumn{2}{c}{(accuracy $\uparrow$)} & \multicolumn{2}{c}{(Lipschitz $\downarrow$)} \\
\cmidrule(lr){2-3} \cmidrule(lr){4-5}
\textbf{Benchmark} & \textbf{1$\times$} & \textbf{4$\times$} & \textbf{Encoder} & \textbf{Decoder} \\
\midrule
Average & 1.0000 & 0.9975 & 24.46 / 11.18 & 1.62 / 4.72 \\
\bottomrule
\end{tabular*}
\vspace{1mm}
\label{tab:textvae}
\end{table}

\begin{table}[t]
\centering
\caption{
\textbf{Self correction performance.}
Natural Language Inference (NLI) measures the semantic consistency between the generated text and the ground-truth text.
Reward Model (RM) evaluates the quality of correction using an external learned reward model
(\texttt{ArmoRM-Llama3-8B-v0.1}).
BERTScore (Score) evaluates semantic similarity, and Levenshtein Distance (Lev.) measures the average edit distance to the ground-truth text.
}
\footnotesize
\setlength{\tabcolsep}{0pt}   %
\begin{tabular*}{\linewidth}{@{\extracolsep{\fill}}lcccc@{}} %
\toprule
\textbf{Model}
& \textbf{NLI} $\uparrow$ & \textbf{RM} $\uparrow$ & \textbf{Score} $\uparrow$ & \textbf{Lev.} $\downarrow$ \\
\midrule
Qwen2.5-1.5B-Instruct & 0.703 & \textbf{1.742} & 0.897 & 173.9 \\
Qwen2.5-1.5B & 0.821 & 0.355 & 0.542 & 1082.1 \\
\midrule
\textbf{Ours-0.6B} & \textbf{0.822} & 1.182 & \textbf{0.899} & \textbf{143.7} \\
\bottomrule
\end{tabular*}
\label{tab:text_correction}
\end{table}

\begin{table}[t]
\centering
\caption{
\textbf{Text infilling quality.} 
MAUVE measures the similarity between model-generated infillings and reference texts, given prefix–suffix pairs.
Higher values indicate better alignment with human-like text distributions.
}
\footnotesize
\setlength{\tabcolsep}{0pt}
\begin{tabular*}{\linewidth}{@{\extracolsep{\fill}}
  >{\centering\arraybackslash}p{0.8\linewidth}   %
  >{\centering\arraybackslash}p{0.2\linewidth}   %
  @{}}
\toprule
\textbf{Model} & \textbf{MAUVE} $\uparrow$ \\
\midrule
Qwen2.5-1.5B-Instruct~\cite{Yang2024Qwen2TR} & 0.569 \\
Qwen2.5-1.5B~\cite{Yang2024Qwen2TR} & 0.543 \\
Qwen2.5-3B~\cite{Yang2024Qwen2TR} & 0.593 \\
Qwen2.5-3B-Instruct~\cite{Yang2024Qwen2TR} & 0.543 \\
\midrule
\textbf{Ours-0.6B} & \textbf{0.616} \\
\bottomrule
\end{tabular*}
\label{tab:infilling_qwen_ours}
\end{table}

\subsection{Evaluation on Global Closed-Loop Generation}
\label{sec:boundary_eval}

While zero-shot perplexity validates density estimation as an Initial Value Problem (IVP), it offers limited insight into global feedback mechanisms. To analyze the \textbf{Closed-Loop Generation} performance of our framework, we extend our evaluation to \textit{Boundary Value Problems (BVPs)}, formulating generation as a trajectory optimization task under constraints. Such tasks are challenging for standard causal AR models due to their unidirectional visibility and for discrete diffusion due to the lack of smooth gradient guidance, yet they naturally fit the operating mode of our optimal control policy.

\noindent
\textbf{Constrained Self-Correction.} 
We first evaluate \emph{error correction}, a task requiring the model to act as a closed-loop controller that projects off-manifold (noisy, hallucination) states back onto the clean data manifold.
We introduce multi-granular perturbations to the WikiText-2 and OpenWebText datasets: (1) \textit{Lexical Noise} (homoglyph/deletion), (2) \textit{Semantic Swaps} (entity substitution), and (3) \textit{Logical Permutations} (temporal reordering).
Standard AR models, constrained by unidirectional attention, treat errors as context to be continued rather than states to be corrected. In contrast, our model utilizes the learned vector field as a \emph{restoring force} to minimize transport energy toward the valid distribution.

In Table \ref{tab:text_correction}, we report NLI consistency and Reward Model (RM) scores to measure semantic validity, alongside Levenshtein Distance to penalize unrestricted rewriting.
As shown in Table \ref{tab:text_correction}, our Manta-LM achieves a favorable balance between high semantic fidelity (0.835 BERTScore) and low edit distance (169.1).
While baselines like Qwen2.5-Instruct~\cite{Yang2024Qwen2TR} achieve high NLI scores by rewriting the entire sentence (high Levenshtein), our model performs \textit{surgical} corrections. This empirically confirms our theoretical claim: the optimal controller is robust in latent control space, effectively self-correction the error states while preserving the uncorrupted global structure.

\noindent
\textbf{Generative Interpolation (i.e., Long-Form Infilling).}
We further task the model with \emph{long-form infilling}, bridging two distant boundary conditions, a capability requiring long-horizon look-ahead.
Using OpenWebText, we construct a challenging benchmark with an aggressive 1:8:1 split (masking the central 80\% of text), stratified across sequence lengths up to 1024 tokens.
This task is challenging for standard ARMs: without specialized Fill-In-the-Middle (FIM) pre-training, causal models do not directly condition on the suffix (future boundary), which can result in incoherent bridges. 

In Table \ref{tab:infilling_qwen_ours}, we prioritize distributional alignment and coherence using the MAUVE score.
Our method demonstrates exceptional infilling performance, successfully generating coherent transitions that respect both prefix and suffix constraints.
This validates that our model solves the \emph{Two-Point Boundary Value Problem} by relaxing the latent trajectory between fixed endpoints. By leveraging the continuous topology of the latent space, our approach avoids the discontinuity issues of discrete infilling, establishing a new paradigm for controllable long-context generation.

\subsection{Model Evaluation and Analysis}

\begin{figure}[t]
    \centering
    \includegraphics[width=0.48\textwidth]{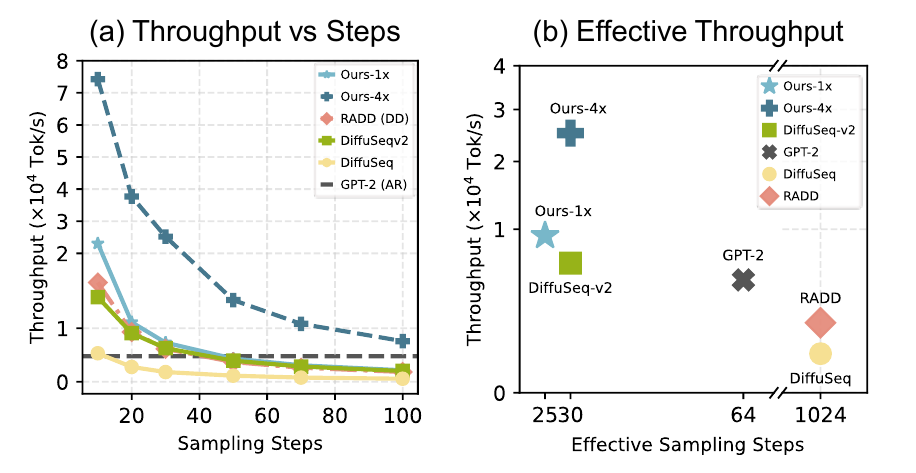}
    \caption{Efficiency evaluation with inference throughput.}
    \vspace{-10pt}
    \label{fig:effi}
\end{figure}

\noindent
\textbf{Efficiency Analysis.}
Figure \ref{fig:effi} demonstrates the dual efficiency gains of our framework. By leveraging $4\times$ latent compression and an energy-minimizing straight trajectory, our model achieves nearly $4\times$ higher throughput per step while requiring significantly fewer sampling steps (NFE) than baselines. Crucially, our inference cost is decoupled from sequence length ($O(1)$ steps), breaking the linear dependency ($O(N)$) inherent to AR models. For 1024-token generation, this culminates in a \textbf{4.6$\times$} speedup over the most efficient baseline, validating the practical scalability of our optimal control formulation.

\noindent
\textbf{Geometric Analysis: Reduce Stiffness.} 
To validate the necessity of the latent space and theoretical claims made in \cref{sec:theory}, we compare the transport dynamics in the raw embedding space versus our rectified latent manifold using two metrics: 
(1) \emph{Trajectory Curvature} $\kappa(t) \approx {\| \ddot{\mathbf{z}}_t \|}/{\| \dot{\mathbf{z}}_t \|^2}$, which quantifies the deviation from the optimal straight-line transport; and
(2) \emph{Vector Field Stiffness} $\mathcal{S}(\mathbf{z}) = \| \nabla_{\mathbf{z}} v_t(\mathbf{z}) \|_F$, which estimates the local Lipschitz constant governing numerical stability.

\begin{figure}[t]
    \centering
    \includegraphics[width=0.48\textwidth]{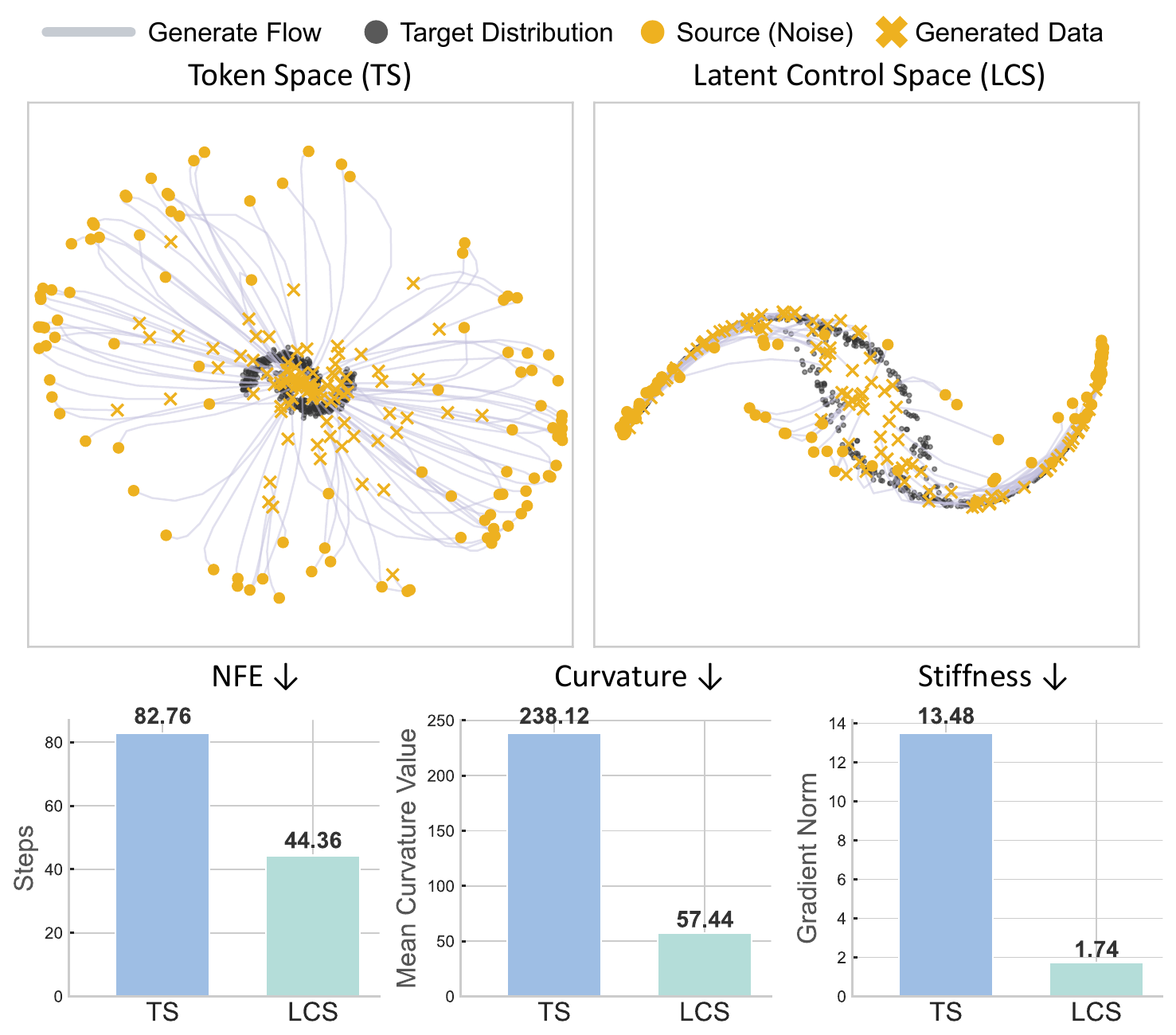}
    \caption{\textbf{Stiffness Analysis.} The raw Token (Embedding) Space exhibits extreme stiffness and high curvature, indicating an ill-conditioned control landscape that forces adaptive solvers (RK45) to high NFE. In contrast, our Rectified Latent Space maintains low stiffness and near-linear trajectories, verifying the efficacy of VAE.}
    \vspace{-10pt}
    \label{fig:stiff}
\end{figure}

\textbf{Ill-Conditioned Embedding vs. Rectified Latent.}
As shown in Figure \ref{fig:stiff}, the raw embedding space exhibits extreme stiffness and curvature. This geometric pathology indicates a non-convex landscape where the vector field must undergo drastic directional changes. According to the Picard-Lindelöf theorem, such exploding Lipschitz constants ($\mathcal{S} \to \infty$) render the ODE \textit{stiff}, forcing adaptive solvers to take infinitesimally small steps (high NFE) to maintain stability.
In contrast, our latent model maintains low stiffness and near-linear trajectories. This empirically confirms that the VAE performs \emph{Manifold Rectification}, transforming the ill-conditioned high-frequency regression problem into a well-conditioned one, thereby making the HJB-inspired dynamics easier to approximate with Flow Matching and efficient large-step integration.

\begin{figure}[t]
    \centering
    \includegraphics[width=0.48\textwidth]{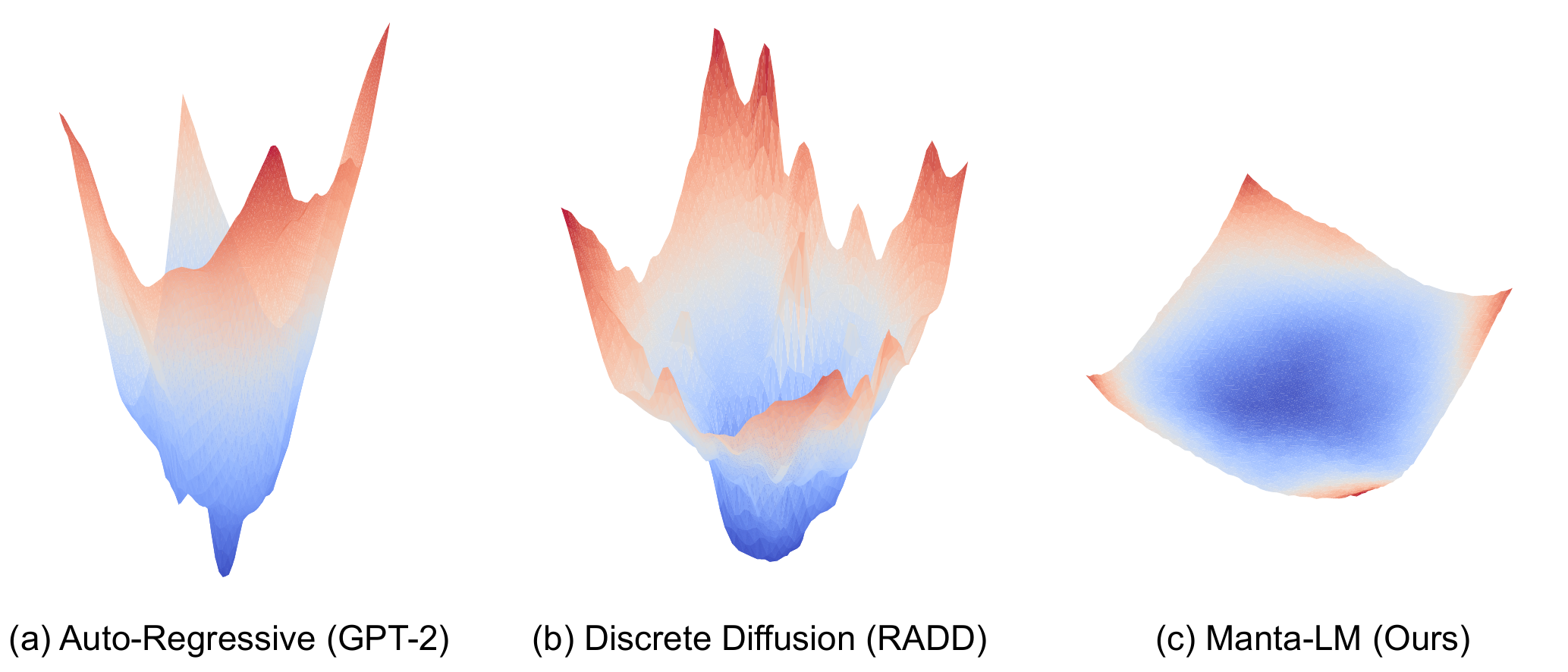}
    \caption{
\textbf{Optimization landscapes of different generation paradigms.}
(a) AR exhibits sharp and unstable geometry. (b) Discrete diffusion leads to fragmented and irregular landscapes. (c) Our Manta-LM  yields a smooth and well-conditioned landscape, enabling stable optimization.
    }
    \vspace{-10pt}
    \label{fig:robustness}
\end{figure}

\noindent
\textbf{Geometric Regularity and Optimization Stability.} 
\cref{fig:robustness} contrasts the rugged optimization landscape of discrete baselines, which reflects severe ill-conditioning, with the smooth and wide-valley geometry induced by our CTRL-LM. This topological regularity empirically supports the effect of manifold rectification. By transforming a chaotic combinatorial search process into a well-conditioned optimal control problem, our method yields intrinsically stable dynamics and demonstrates clear advantages over brittle discrete formulations.

\section{Conclusion}
We presented \textbf{Manta-LM}, a framework that studies and re-imagines text generation as Stochastic Optimal Control problem. By approximating HJB-inspired dynamics with Flow Matching on a rectified manifold, our model overcomes the ``myopic'' limitations of autoregressive baselines, enabling global trajectory planning and closed-loop feedback-based refinement. Empirical results show strong zero-shot density estimation and favorable performance on Boundary Value Problems, such as non-causal infilling and error correction, that are difficult for standard causal models without specialized mechanisms. This work connects discrete language modeling with continuous dynamical systems and offers a robust, mathematically grounded path toward efficient and controllable text generation.

\paragraph{Potential Limitations and Future Directions}
While the proposed global planning paradigm enables stable and well-guided generation, it may be computationally less efficient for interruptible agentic tasks, such as tool calling, compared to fine-grained autoregressive inference. Adapting continuous flow dynamics to support low-cost, sequential interruptions without incurring redundant computation remains an important direction for future work.

\section*{Impact Statement}
This paper presents work whose goal is to advance the field of machine learning. There are many potential societal consequences of our work, none of which we feel must be specifically highlighted here.

\bibliography{example_paper}
\bibliographystyle{icml2026}

\newpage
\appendix
\onecolumn
\section{Appendix}
\subsection{Find the Optimal Control Law via Solving HJB}
We formally derive the optimal control law that governs our Latent Flow LLM. Following the Benamou-Brenier formulation for dynamic optimal transport \cite{benamou2000computational}, we seek a control policy $\mathbf{u}^*$ that minimizes the transport cost functional $J(\mathbf{u})$.
In the specific context of text generation, the \textit{Terminal Cost} $\Phi$ is strictly defined by the data likelihood:
\begin{equation}
    \Phi(\mathbf{z}_1) = -\log p_{\text{data}}(\mathbf{z}_1).
\end{equation}
Thus, the total objective functional becomes:
\begin{equation}
    J(\mathbf{u}) = \mathbb{E} \left[ \underbrace{-\log p_{\text{data}}(\mathcal{D}(\mathbf{z}_1))}_{\text{Semantic Fidelity}} + \lambda \int_0^1 \underbrace{\frac{1}{2} \|\mathbf{u}_t(\mathbf{z}_t)\|^2}_{\text{Transport Energy}} dt \right].
    \label{eq:control_objective}
\end{equation}

\textbf{The Hamilton-Jacobi-Bellman (HJB) Equation.}
To solve for the optimal trajectory, we define the \textit{Value Function} $V(\mathbf{z}, t)$ as the minimum expected cost-to-go from state $\mathbf{z}$ at time $t$:
\begin{equation}
    V(\mathbf{z}, t) = \inf_{\mathbf{u}} \mathbb{E} \left[ \Phi(\mathbf{z}_1) + \int_t^1 \frac{1}{2} \|\mathbf{u}\|^2 d\tau \bigg| \mathbf{z}_t = \mathbf{z} \right].
\end{equation}
According to the \textit{Dynamic Programming Principle} in continuous time, the value function $V$ must satisfy the Hamilton-Jacobi-Bellman (HJB) partial differential equation:
\begin{equation}
    -\frac{\partial V}{\partial t} = \inf_{\mathbf{u}} \mathcal{H}(\mathbf{z}, \mathbf{u}, \nabla_{\mathbf{z}} V),
    \label{eq:hjb_pde}
\end{equation}
where the Hamiltonian $\mathcal{H}$ represents the total energy of the system:
\begin{equation}
    \mathcal{H}(\mathbf{z}, \mathbf{u}, \nabla_{\mathbf{z}} V) = \underbrace{\frac{1}{2} \|\mathbf{u}\|^2}_{\text{Kinetic}} + \underbrace{\mathbf{u} \cdot \nabla_{\mathbf{z}} V}_{\text{Potential Interaction}}.
\end{equation}

\textbf{The Optimal Control Law.}
The optimal control $\mathbf{u}^*$ is obtained by minimizing the Hamiltonian with respect to $\mathbf{u}$. Setting the gradient $\nabla_{\mathbf{u}} \mathcal{H} = 0$, we derive the analytical form of the optimal controller:
\begin{equation}
    \mathbf{u} + \nabla_{\mathbf{z}} V = 0 \implies \mathbf{u}^*(\mathbf{z}, t) = -\nabla_{\mathbf{z}} V(\mathbf{z}, t).
    \label{eq:optimal_u}
\end{equation}

\textbf{Manifestation in Text Generation: The Semantic Gradient Flow.}
Equation (\ref{eq:optimal_u}) reveals the physical essence of our generation process. In the context of LLMs, the value function $V(\mathbf{z}, t)$ acts as a time-varying \textbf{Semantic Potential Field}:
\begin{itemize}
    \item \textbf{Potential Landscape:} High values of $V$ correspond to regions of low semantic coherence or syntactic errors; low values correspond to the data manifold (valid text).
    \item \textbf{Gradient Guidance:} The optimal control law $\mathbf{u}^* = -\nabla_{\mathbf{z}} V$ dictates that the latent state should strictly follow the direction of \textit{steepest semantic descent}.
\end{itemize}
Consequently, our Manta-LM functions as an \textbf{Optimal Closed-Loop Controller}. Unlike AR models that predict the next token from history alone, our controller uses the global gradient of the semantic potential field ($-\nabla V$), steering the flow to actively minimize the discrepancy between the current latent state and valid target text.

\subsection{Language Models as Sub-optimal Controller}
\label{sec:appendix_pathologies}

Under the continuous-control view, AR and Discrete Diffusion from the optimal control law $\mathbf{u}^* = -\nabla V$ are not merely architectural choices but fundamental violations of the well-posedness conditions for dynamical systems. We analyze three resulting issues: \textit{Energy Divergence}, \textit{Adjoint State Vanishing}, and \textit{Gradient Absence}.

\paragraph{1. Trajectory Singularity and System Stiffness.}
Optimal transport mandates minimizing the kinetic energy action $\mathcal{A} = \frac{1}{2} \int_0^1 \|\mathbf{u}_t\|^2 dt$.
\begin{itemize}
    \item \textbf{Impulsive AR dynamics:} For Autoregressive models, substituting the impulsive control law (Eq. \ref{eq:ar_sde}) into the action functional yields a divergence:
    \begin{equation}
        \mathcal{A}_{\text{AR}} \propto \int_0^1 \left\| \sum \mathbf{f}_k \delta(t-t_k) \right\|^2 dt \to \infty.
    \end{equation}
    Since the $L^2$ norm of a Dirac delta is undefined (infinite energy), the trajectory exhibits \textbf{Infinite Instantaneous Curvature}. In numerical analysis, this characterizes a highly stiff system ($\lambda_{\max} \to \infty$), making it poorly suited to parallel ODE solvers and reflecting the serial nature of AR integration.
    
    \item \textbf{Lipschitz Explosion:} For continuous diffusion on raw embeddings, the irregular topology implies that the gradient field $\nabla \log p_t$ is not Lipschitz continuous. The local Lipschitz constant $L$ explodes:
    \begin{equation}
        L = \sup_{\mathbf{z} \neq \mathbf{y}} \frac{\| \mathbf{u}(\mathbf{z}) - \mathbf{u}(\mathbf{y}) \|}{\| \mathbf{z} - \mathbf{y} \|} \gg 1.
    \end{equation}
    Stability of numerical integration requires step sizes $\Delta t < 2/L$. As $L \to \infty$, $\Delta t \to 0$, causing the Number of Function Evaluations (NFE) to diverge as shown in \cref{fig:stiff}.
\end{itemize}
This stiffness helps explain the serial or high-NFE behavior observed in these settings: highly non-smooth dynamics (singular) are difficult for parallel continuous solvers (like Picard iteration) to integrate efficiently.

\paragraph{2. Open-Loop Drift via Adjoint State Vanishing.}
In control theory, the optimality of a decision at time $t$ is determined by the \textit{Prediction Horizon} $H$. A general Model Predictive Control (MPC) objective minimizes the cost-to-go over the interval $[t, t+H]$:
\begin{equation}
    J_H(\mathbf{u}_t) = \mathbb{E} \left[ \int_t^{t+H} \frac{1}{2} \|\mathbf{u}_\tau\|^2 d\tau + \Phi(\mathbf{z}_{t+H}) \right].
\end{equation}
where $\Phi(\mathbf{z}_{t+H})$ denotes the \textbf{Terminal Potential}, which encodes the global semantic coherence and validity of the final outcome.
In optimal control, trajectory stability is governed by the \textit{Adjoint Equation} for the co-state $\mathbf{p}_t$ (the sensitivity of the terminal cost $\Phi$):
\begin{equation}
    \dot{\mathbf{p}}_t = -\nabla_{\mathbf{z}} \mathcal{H}(\mathbf{z}_t, \mathbf{u}_t, \mathbf{p}_t), \quad \mathbf{p}_1 = -\nabla_{\mathbf{z}} \Phi(\mathbf{z}_1).
\end{equation}
\begin{itemize}
    \item \textbf{Adjoint Vanishing:} Global optimality requires $H = 1-t$ (solving until the terminal state). We characterize AR generation as a degenerate MPC policy where the horizon is truncated to a single discrete step with horizon $H=1$. It truncates the backward pass, effectively forcing $\mathbf{p}_t \equiv \mathbf{0}$ for $t < 1$.
    \item \textbf{Lyapunov Instability:} Without the restoring force provided by the adjoint feedback $\mathbf{u}_{\text{fb}} \propto \mathbf{p}_t$, the error dynamics $\mathbf{e}_t = \mathbf{z}_t - \mathbf{z}^*_t$ become unstable. Any perturbation $\epsilon$ (e.g., quantization noise) grows exponentially, leading to irreversible drift (hallucination):
    \begin{equation}
        \frac{d}{dt} \|\mathbf{e}_t\| > 0 \quad (\text{Open-Loop Instability}).
    \end{equation}
    Geometrically, this resembles a ball rolling along a narrow ridge without a restoring force—any deviation pushes it irreversibly off the data manifold (hallucination).
\end{itemize}

\paragraph{3. Geometric Blindness via Metric Singularity.}
Efficient optimization requires a \textit{Gradient Flow} on a Riemannian manifold $(\mathcal{M}, g)$.
\begin{itemize}
    \item \textbf{Gradient Absence:} 
    To clarify the absence of a smooth gradient on the discrete lattice $\mathcal{D}$, we invoke the definition of the Fréchet derivative on Riemannian manifolds. On a smooth manifold, the gradient $\nabla V(\mathbf{z}) \in T_{\mathbf{z}}\mathcal{M}$ is the unique vector satisfying the linearization condition $\lim_{\|\mathbf{d}\| \to 0} \|V(\mathbf{z}+\mathbf{d}) - V(\mathbf{z}) - \langle \nabla V, \mathbf{d} \rangle\| / \|\mathbf{d}\| = 0$. However, $\mathcal{D}$ possesses the discrete topology where the metric is lower-bounded by $\|\mathbf{y} - \mathbf{z}\| \ge 1$ for $\mathbf{y} \neq \mathbf{z}$, precluding infinitesimal displacements. Thus, the smooth differential operator required by our HJB formulation is not directly available on $\mathcal{D}$. Furthermore, due to the high-frequency discontinuities of the semantic energy landscape $V$ across discrete tokens, no single vector $\mathbf{v}$ can satisfy the first-order Taylor approximation for the local neighborhood, formally implying that $\nexists \ \mathbf{v}$ such that $V(\mathbf{z}+\mathbf{d}) - V(\mathbf{z}) \approx \langle \mathbf{v}, \mathbf{d} \rangle$ holds, thereby confirming the structural absence of gradient guidance.
    \begin{equation}
        \nexists \ \mathbf{v} \in T_{\mathbf{z}}\mathcal{M} \quad \text{s.t.} \quad \langle \mathbf{v}, \mathbf{d} \rangle \approx V(\mathbf{z}+\mathbf{d}) - V(\mathbf{z}).
    \end{equation}
    \item \textbf{Combinatorial Fallback:} Without a smooth descent direction $-\nabla V$, the system relies on stochastic transition search. This degrades the convergence rate from linear/superlinear (Gradient Descent) to sublinear (Random Walk), manifesting as the efficiency-quality trade-off.
\end{itemize}

\subsection{Discussion}
\label{sec:flow_training}

\begin{figure*}[t]
\centering
\includegraphics[width=\linewidth]{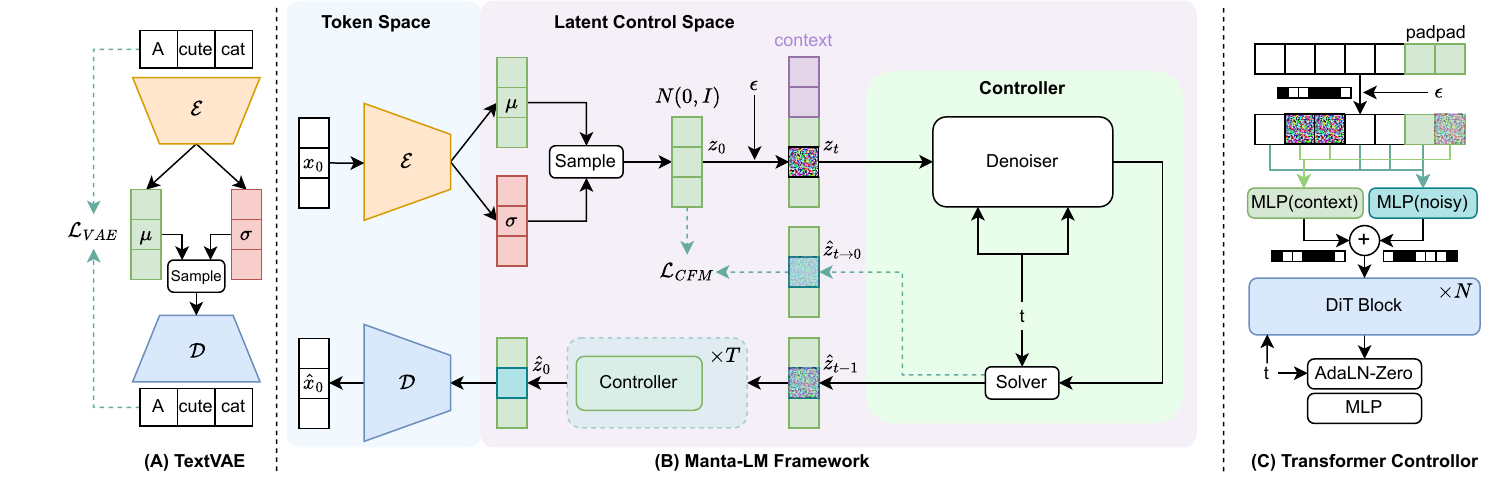}
\caption{Model structure and pipeline. }
\vspace{-8pt}
\label{fig:model}
\end{figure*}

Functioning as an optimal controller, our framework not only subsumes the capabilities of standard AR models but also unlocks some entirely novel tasks.

\subsubsection{Universal Generation via Boundary Value Problems}
Unlike Autoregressive models, which are architecturally constrained to causal generation (Prefix $\to$ Suffix), our flow-based formulation treats text generation as a generic \textbf{Boundary Value Problem (BVP)}.
In our control framework, different generation tasks (e.g., continuation, in-filling) correspond to solving the same ODE but with different Boundary Conditions.
We define a masking operator $\mathcal{M}$ that partitions the latent state into a "Constraint" region $\mathbf{z}_{\text{ctx}}$ (fixed boundary condition) and a "State" region $\mathbf{z}_{\text{flow}}$ (free variable).
The vector field is then conditioned on the constraints:
\begin{equation}
    \frac{d\mathbf{z}_{\text{flow}}}{dt} = v_\theta(\mathbf{z}_{\text{flow}} \cup \mathbf{z}_{\text{ctx}}, t).
\end{equation}

During training, we stochastically simulate diverse boundary conditions (Causal, In-filling, Prefix) as shown in \cref{fig:model}. This unifies all text generation tasks into a single mathematical formulation: \textit{trajectory optimization under partial state constraints}.

\subsubsection{Topological Padding for Variable Lengths}
To handle variable sequence lengths within this rigid ODE framework, we introduce a learnable \texttt{<PAD>} token embedding to represent the \textbf{"Null State"} (a region of zero potential and void semantics).
We imply a "Topological Boundary" condition where the flow must converge to the Null Attractor (Null State) in padding regions. By training the model to flow from noise to \texttt{<PAD>} in these areas as shown in \cref{fig:model}, the controller learns the manifold's boundary. During inference, this allows the ODE solver to dynamically determine the sequence length by identifying when the flow state converges to the Null attractor.

\subsection{Latent Control Space and TextVAE Evaluation.} 
As the core module to build the control-friendly latent control space, TextVAE's reconstruction fidelity and robustness are critical to the performance of text generation.
Our TextVAE trained on a diverse dataset to ensure broad generalization, covering \textbf{Code} (OpenCoder\cite{huang2025opencoder}, Infinity-Instruct\cite{li2025infinity}), \textbf{Mathematics} (OpenMath\cite{moshkov2025aimo}, GSM8K\cite{cobbe2021gsm8k}), and \textbf{General NLP} (Wiki, Common Crawl, LM1B\cite{Chelba2013OneBW}).

\subsubsection{TextVAE Architecture and Objective}
The TextVAE follows the design spirit of Deep Compression Autoencoders~\cite{chen2024deep}, adapted from 2D images to 1D token sequences. The encoder and decoder are symmetric, fully parallel convolutional networks. We do not use self-attention inside the VAE: each latent is computed from a local window of tokens, which preserves token-to-latent alignment and avoids global cross-position entanglement before the diffusion controller.

The VAE objective combines reconstruction, posterior regularization, and feature-stability terms:
\begin{equation}
\mathcal{L}_{\mathrm{VAE}}
= \mathcal{L}_{\mathrm{rec}}
+ \beta(t)\mathcal{L}_{\mathrm{KL}}
+ \lambda_{\mathrm{stab}}\mathcal{L}_{\mathrm{stab}}.
\end{equation}
For $\mathcal{L}_{\mathrm{rec}}$, the decoder first maps latent states to continuous token embeddings. Vocabulary logits are then computed by dot product with the shared embedding matrix, followed by learnable scale and bias terms. Binary cross-entropy is applied between these logits and one-hot token targets; empirically, this converged faster and gave better reconstruction than standard cross-entropy in our setting. The KL term regularizes the Gaussian posterior $q_\phi(\mathbf{z}\mid x)$ toward $\mathcal{N}(0,I)$ with a progressive schedule $\beta(t)$ to reduce posterior collapse. The stability loss regularizes intermediate features and improves numerical robustness under latent perturbations.

Decoding is fully parallel. Given a latent sequence, the convolutional decoder outputs continuous embeddings at all positions, projects them to vocabulary logits using the shared token embedding matrix with learnable scale/bias, applies softmax, and removes padding tokens when producing the final text. No autoregressive dependency is used in the VAE decoder.

\subsubsection{Local Integral Operators and Downsampling}
The convolutional blocks in TextVAE can be viewed as Local Integral Operators: the output at each latent position aggregates information from a finite local neighborhood through a translation-equivariant kernel. This locality is important for structured generation because token-level composition, such as concatenating constrained and free segments, remains reflected in the latent representation.

Downsampling is configurable. In the $4\times$ setting used in our main experiments, the encoder applies two downsampling stages and maps a length-$L$ token sequence to a length-$L/4$ latent sequence. Downsampling reduces sequence length while preserving local order and approximate local token-to-latent alignment, rather than collapsing the sentence into a globally mixed representation.

For boundary-conditioned generation, we first group tokens according to the VAE compression factor and pad within each group as needed. For example, a constraint pattern \texttt{C\_\_CCC\_\_\_\_}, where \texttt{C} denotes constrained tokens and underscores denote free slots, becomes \texttt{|C000|\_\_00|CCC0|\_\_\_\_|} under $4\times$ grouping, where \texttt{0} denotes \texttt{<PAD>}. The convolutional encoder maps each group to its corresponding latent position. Groups containing constrained tokens are encoded as context latents, while fully unconstrained groups are initialized from Gaussian noise. We use separate embeddings to mark context and free latents before passing them to the global controller, and the decoder removes \texttt{<PAD>} tokens after generation. This procedure supports both initial value problems (prefix-only constraints) and boundary value problems (prefix, suffix, or arbitrary constrained spans).

\begin{table}[h]
\centering
\small
\setlength{\tabcolsep}{6pt}
\begin{tabular}{lcc}
\toprule
\textbf{Benchmark} & \textbf{\begin{tabular}[c]{@{}c@{}}Reconstruction\\(accuracy $\uparrow$)\end{tabular}} & \textbf{\begin{tabular}[c]{@{}c@{}}Robustness\\(Lipschitz $\downarrow$)\end{tabular}} \\
\midrule
HumanEval & 1.0000 / 0.9954 & \multirow{20}{*}{\begin{tabular}[c]{@{}c@{}}
Encoder:\\ 24.46 / 11.18 \\[4pt] %
Decoder:\\ 1.62 / 4.72
\end{tabular}} \\
MBPP & 1.0000 / 0.9980 &  \\
BBH & 1.0000 / 0.9973 &  \\
ARC & 1.0000 / 0.9989 &  \\
TruthfulQA & 1.0000 / 0.9973 &  \\
Winogrande & 1.0000 / 0.9996 &  \\
HellaSwag & 1.0000 / 0.9960 &  \\
PIQA & 1.0000 / 0.9988 &  \\
CMMLU & 1.0000 / 0.9983 &  \\
CEval & 1.0000 / 0.9977 &  \\
MMLU & 1.0000 / 0.9935 &  \\
GSM8K & 1.0000 / 0.9968 &  \\
Math & 1.0000 / 0.9967 &  \\
LM1B & 1.0000 / 0.9997 &  \\
CodeAlpaca & 1.0000 / 0.9990 &  \\
QQP & 1.0000 / 0.9977 &  \\
Wiki & 1.0000 / 0.9962 &  \\
Common Crawl & 1.0000 / 0.9973 &  \\
Quasar-T & 1.0000 / 0.9985 &  \\
\bottomrule
\end{tabular}
\vspace{1mm}
\caption{
\textbf{Reconstruction and robustness of the TextVAE across heterogeneous benchmarks.}
We report token-level reconstruction accuracy at $1\times / 4\times$ latent compression rates.
Robustness is measured by the estimated Lipschitz constants of the encoder and decoder, reflecting the smoothness of the learned latent manifold.
Results demonstrate that our VAE achieves near-lossless compression while maintaining geometric stability.
}
\label{tab:textvae_recon_robust_singlecol}
\end{table}

We evaluate performance across 19 benchmarks focusing on two metrics: (1) \textbf{Reconstruction Accuracy} at $1\times$ and $4\times$ compression; and (2) \textbf{Robustness}, measured by the Lipschitz constant ($\sigma=0.01$).
As shown in Table~\ref{tab:textvae_recon_robust_singlecol}, TextVAE achieves \textbf{lossless reconstruction (1.00)} at $1\times$ and maintains $>99.3\%$ accuracy at $4\times$.
Furthermore, low Lipschitz constants indicate a smooth latent space, which is critical for ensuring that perturbations during diffusion denoising yield semantically consistent decodings.

\subsection{Additional Evaluation Details}

\subsubsection{Continuous Diffusion on Zero-Shot PPL}
Continuous diffusion baselines are primarily trained and evaluated under conditional generation protocols. When evaluated on the unconditional zero-shot PPL setting used in \cref{tab:zero_shot}, they underperform AR and discrete diffusion models because they are not optimized to model the unconditional data distribution. We therefore report them separately here and keep the main comparison aligned with the established discrete diffusion protocol.

\begin{table}[h]
\centering
\caption{Zero-shot perplexity ($\downarrow$) of continuous diffusion baselines under the unconditional density-estimation setting.}
\scriptsize
\setlength{\tabcolsep}{2pt}
\begin{tabular*}{0.72\linewidth}{@{\extracolsep{\fill}}lccccc@{}}
\toprule
\textbf{Method} & \textbf{Lambada} & \textbf{Wikitext2} & \textbf{PTB} & \textbf{Wikitext103} & \textbf{1BW} \\
\midrule
DiffuSeq & 148.63 & 115.34 & 378.42 & 122.87 & 164.29 \\
SeqDiffuSeq & 112.38 & 88.75 & 291.56 & 94.10 & 129.85 \\
\bottomrule
\end{tabular*}
\label{tab:continuous_zero_shot}
\end{table}

\subsubsection{Training and Inference Cost}
Unless otherwise specified, inference measurements use a single NVIDIA RTX A6000 with batch size 64 and sequence length 128.

\begin{table}[h]
\centering
\caption{Practical training and inference cost of Manta-LM.}
\scriptsize
\setlength{\tabcolsep}{3pt}
\begin{tabular*}{0.86\linewidth}{@{\extracolsep{\fill}}lcccc@{}}
\toprule
\textbf{Training setup} & \textbf{Training cost} & \textbf{Infer. memory} & \textbf{Latency/sample} & \textbf{Latency/step} \\
\midrule
80k steps, 2$\times$ A6000 & 658.6 GPU h & 4.2 GB & 29.33 ms & 1.47 ms \\
\bottomrule
\end{tabular*}
\label{tab:cost}
\end{table}
\begin{figure*}[t]
\centering
\includegraphics[width=\linewidth]{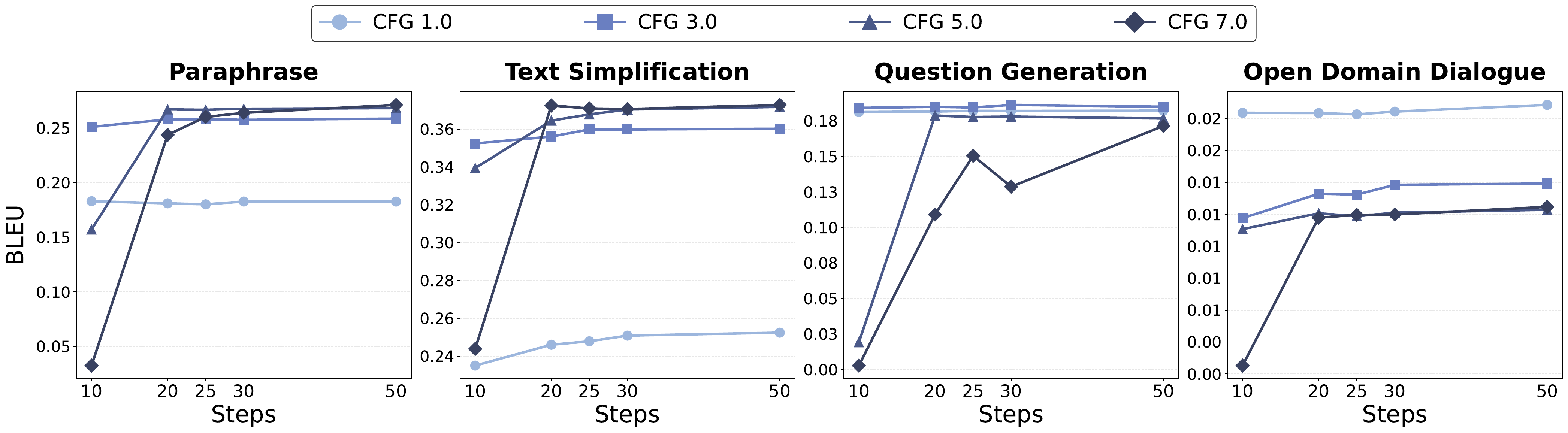}
\caption{Analysis on interplay between CFG guidance strength and integration fidelity}
\vspace{-8pt}
\label{fig:cfg}
\end{figure*}

\subsection{Interplay between Guidance Strength and Integration Fidelity.}
We investigate the joint impact of Classifier-Free Guidance (CFG) strength $w$ and sampling steps $N$ on generation quality in \cref{fig:cfg}. Our analysis reveals three distinct geometric regimes:

\begin{itemize}
    \item \textbf{Under-Guided Regime ($w=1.0$):} Performance is stable invariant to $N$ but consistently suboptimal. The latent flow is smooth but lacks sufficient conditional force to drive the trajectory toward high-fidelity regions.
    \item \textbf{Optimal Regime ($w \in [3.0, 5.0]$):} This setting achieves the best quality-efficiency trade-off. Metrics saturate rapidly (within 20--30 steps), indicating that the vector field is sufficiently aligned with the condition while remaining smooth enough for coarse-step integration.
    \item \textbf{Over-Guided Regime ($w \ge 7.0$):} We observe a sharp performance collapse at low $N$. While quality recovers with finer discretization, this sensitivity highlights a \textit{Stiffness Constraints}.
\end{itemize}

\noindent \textbf{Geometric Interpretation.}
Unlike AR models where guidance acts as a logit bias, in Latent Optimal Control, CFG directly reshapes the velocity field: $v_{w} = v_{\text{uncond}} + w(v_{\text{cond}} - v_{\text{uncond}})$. Increasing $w$ amplifies the conditional gradient, effectively increasing the local Lipschitz constant (curvature) of the dynamics. Strong guidance renders the ODE \textbf{stiff}, necessitating fine-grained temporal discretization (high $N$) to avoid numerical divergence. Consequently, practical deployment requires balancing conditional strength with integration stability, favoring the moderate regime.

\begin{figure*}[t]
\centering
\includegraphics[width=\linewidth]{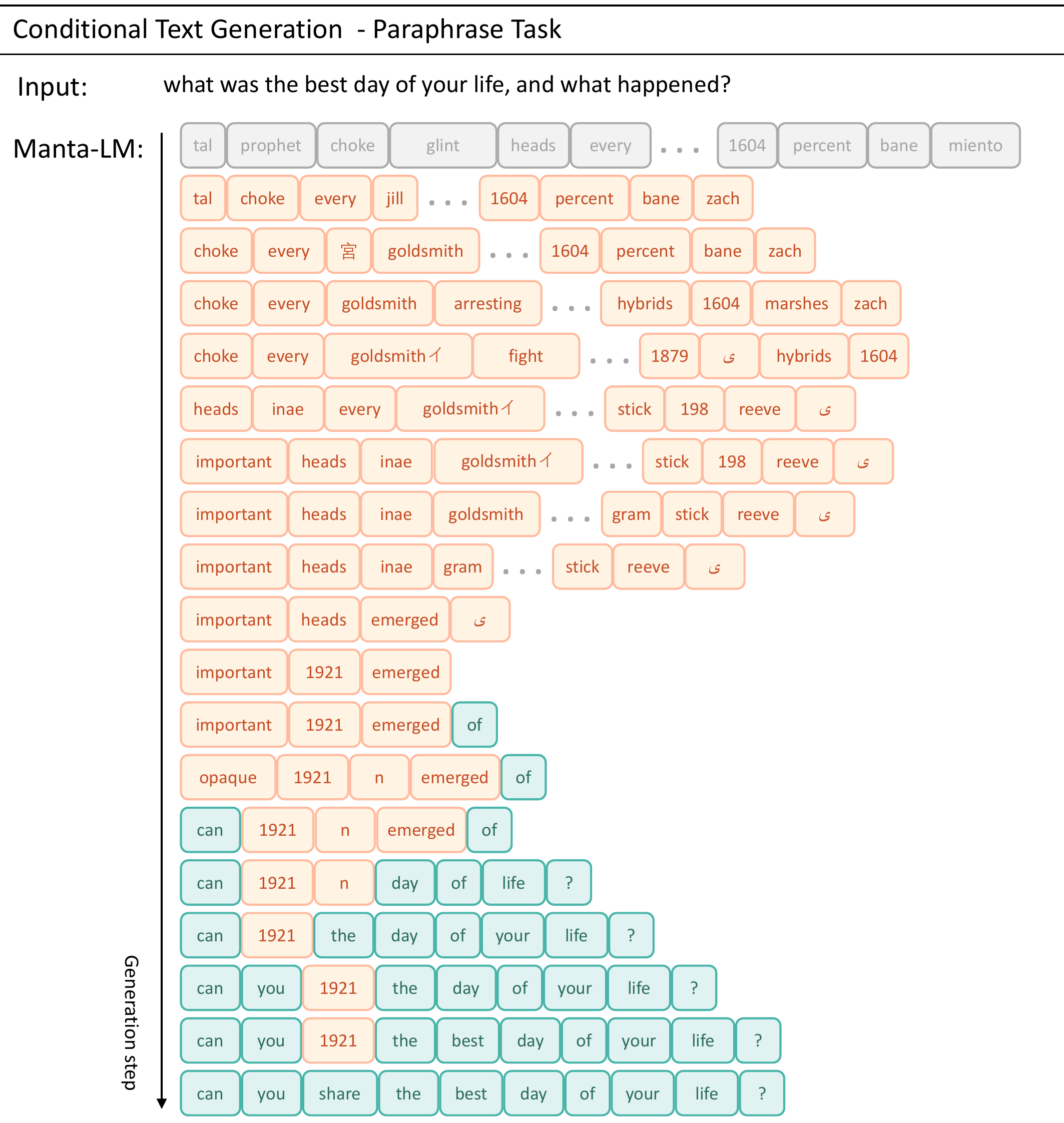}
\caption{Step-by-step conditional generation process of Manta-LM on a paraphrase task.
Given the input sentence ``what was the best day of your life, and what happened?”, the figure visualizes the intermediate generation trajectories of Manta-LM across diffusion steps. }
\vspace{-8pt}
\label{fig:paraphrase_life}
\end{figure*}

\begin{figure*}[t]
\centering
\includegraphics[width=\linewidth]{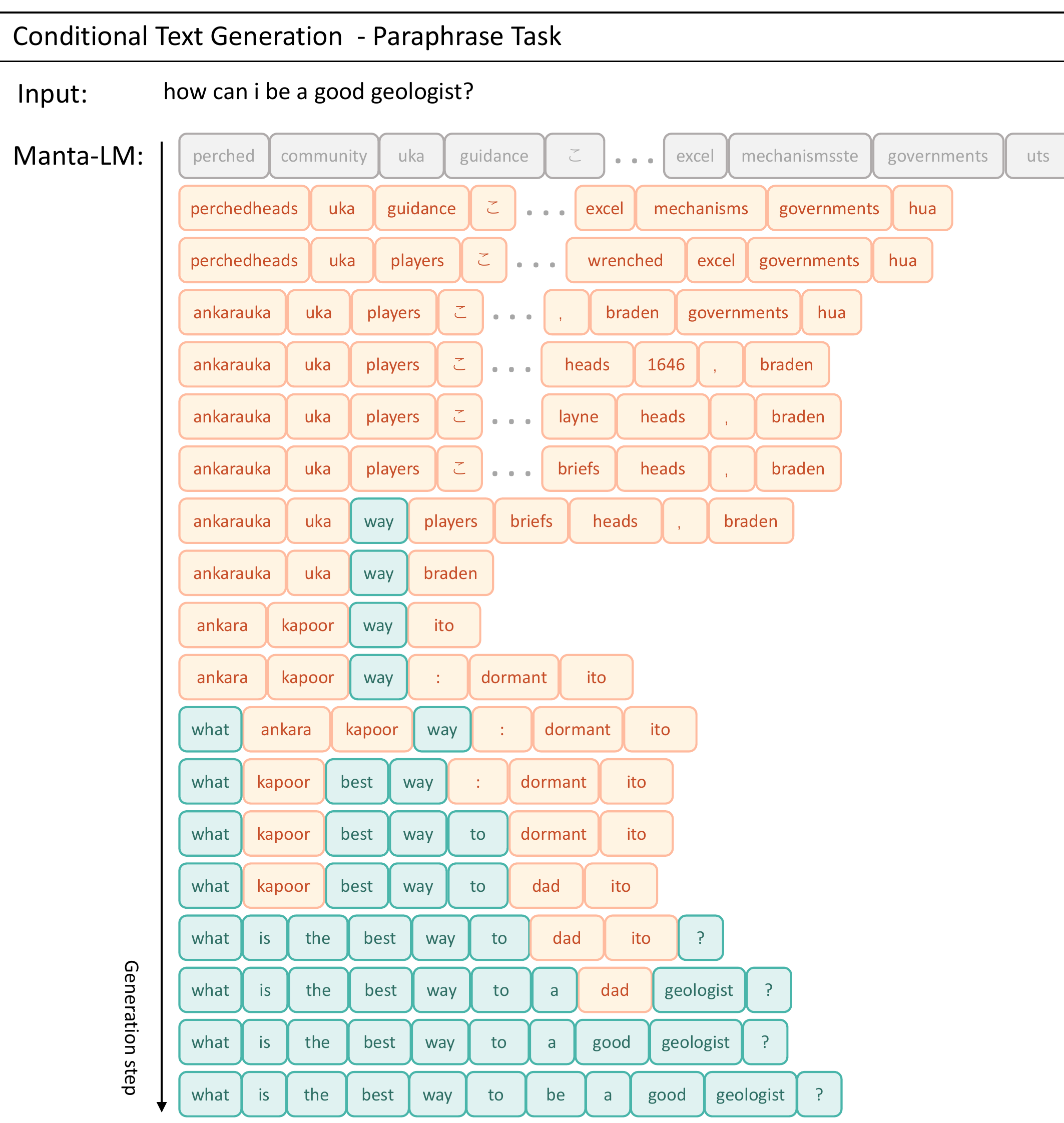}
\caption{Step-by-step conditional generation process of Manta-LM on a paraphrase task.
Given the input sentence ``how can i be a good geologist?”, the figure visualizes the intermediate generation trajectories of Manta-LM across diffusion steps. }
\vspace{-8pt}
\label{fig:paraphrase_geologist}
\end{figure*}

\begin{figure*}[t]
\centering
\includegraphics[width=\linewidth]{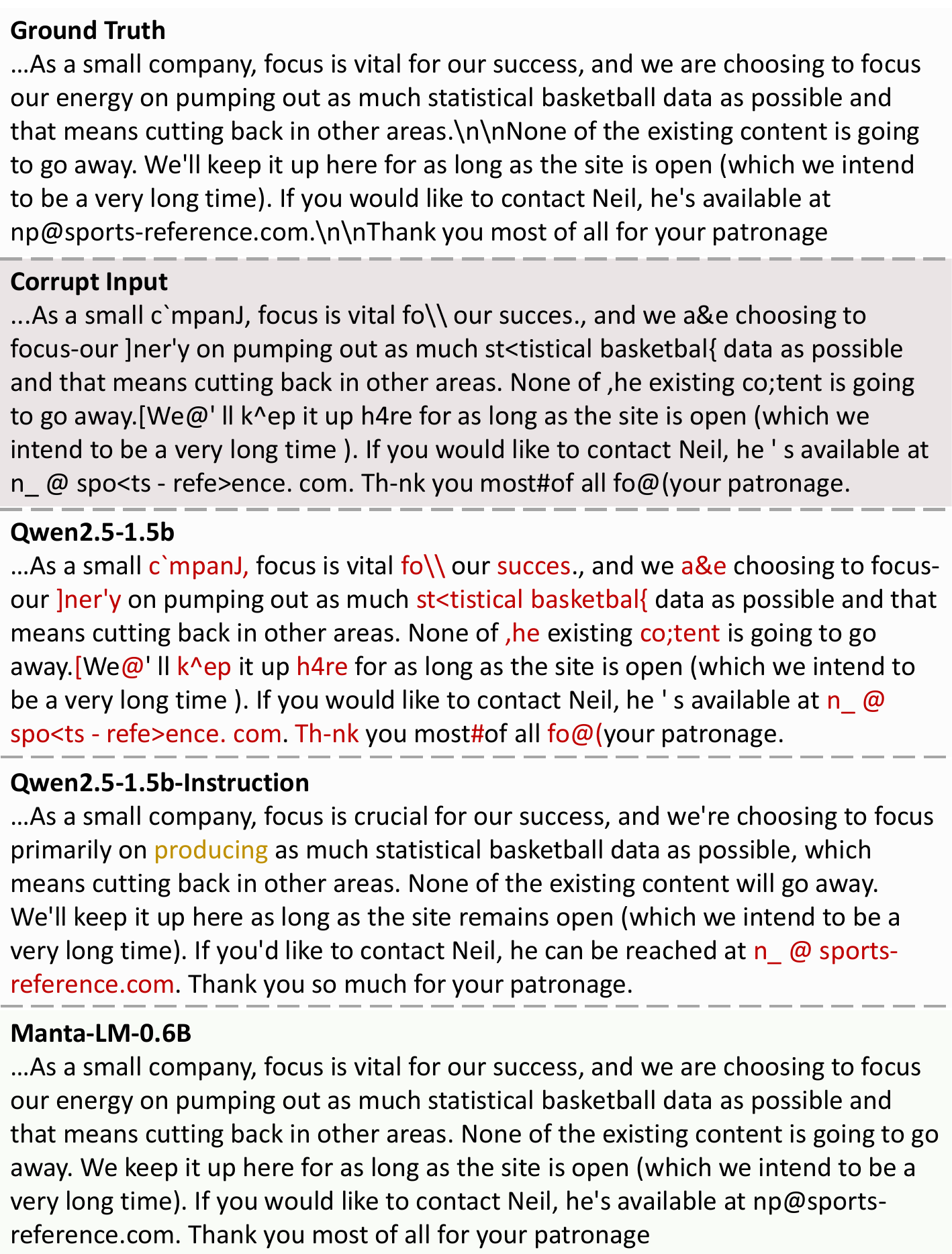}
\caption{Visualizing error correction capabilities across different models.
Red text indicates corrupted or erroneous tokens introduced by noise. 
while yellow text denotes tokens that are semantically consistent with the ground-truth text but differ in surface form.
}
\vspace{-8pt}
\label{fig:correction_examples}
\end{figure*}

\begin{figure*}[t]
\centering
\includegraphics[width=\linewidth]{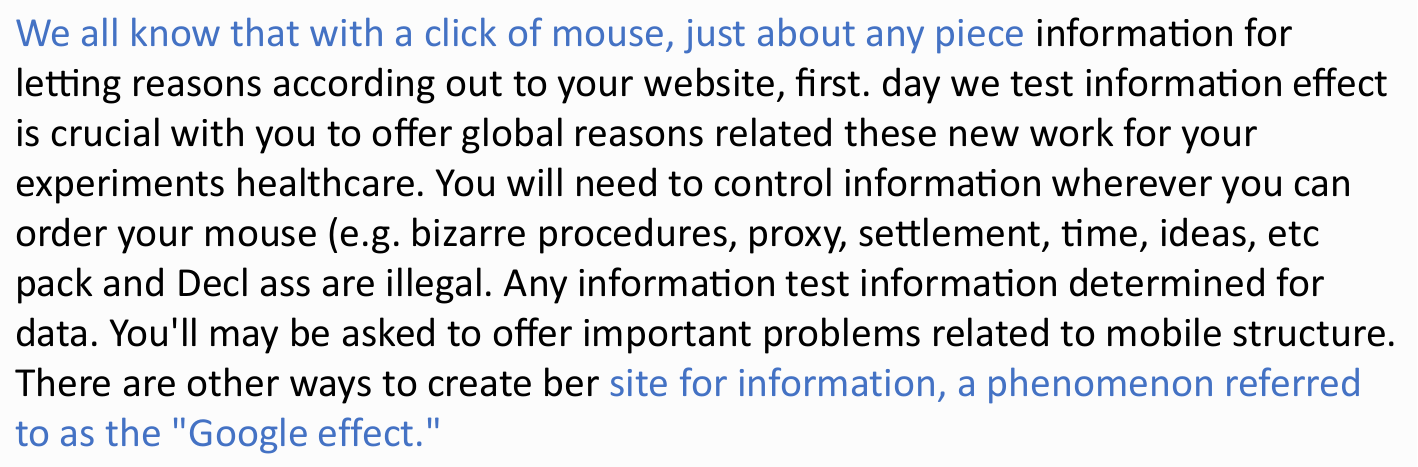}
\caption{Qualitative examples of the text infilling task. Text in {\color{blue}blue} represents the provided prefix and suffix, while text in {\color{black}black} denotes the model's generated results.}
\vspace{-8pt}
\label{fig:infilling_examples}
\end{figure*}

\begin{figure*}[t]
\centering
\includegraphics[width=\linewidth]{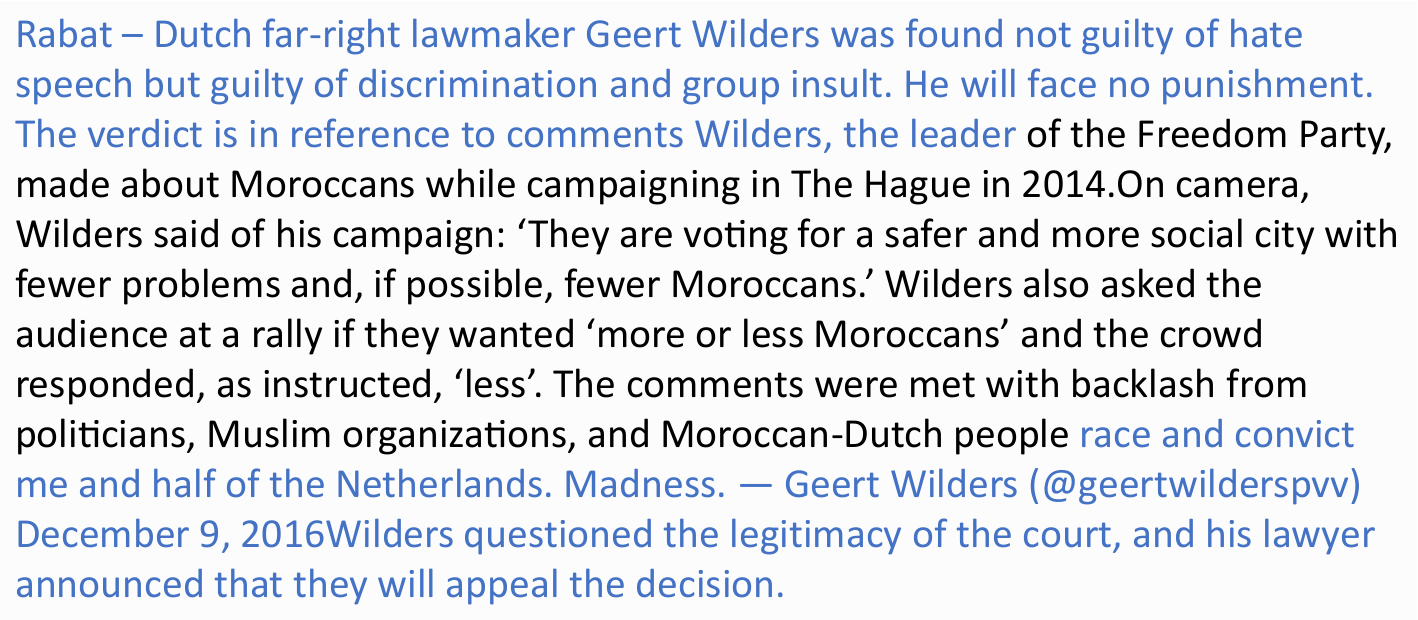}
\caption{Qualitative examples of the text infilling task. Text in {\color{blue}blue} represents the provided prefix and suffix, while text in {\color{black}black} denotes the model's generated results.}
\vspace{-8pt}
\label{fig:infilling_examples_v1}
\end{figure*}

\begin{figure*}[t]
\centering
\includegraphics[width=\linewidth]{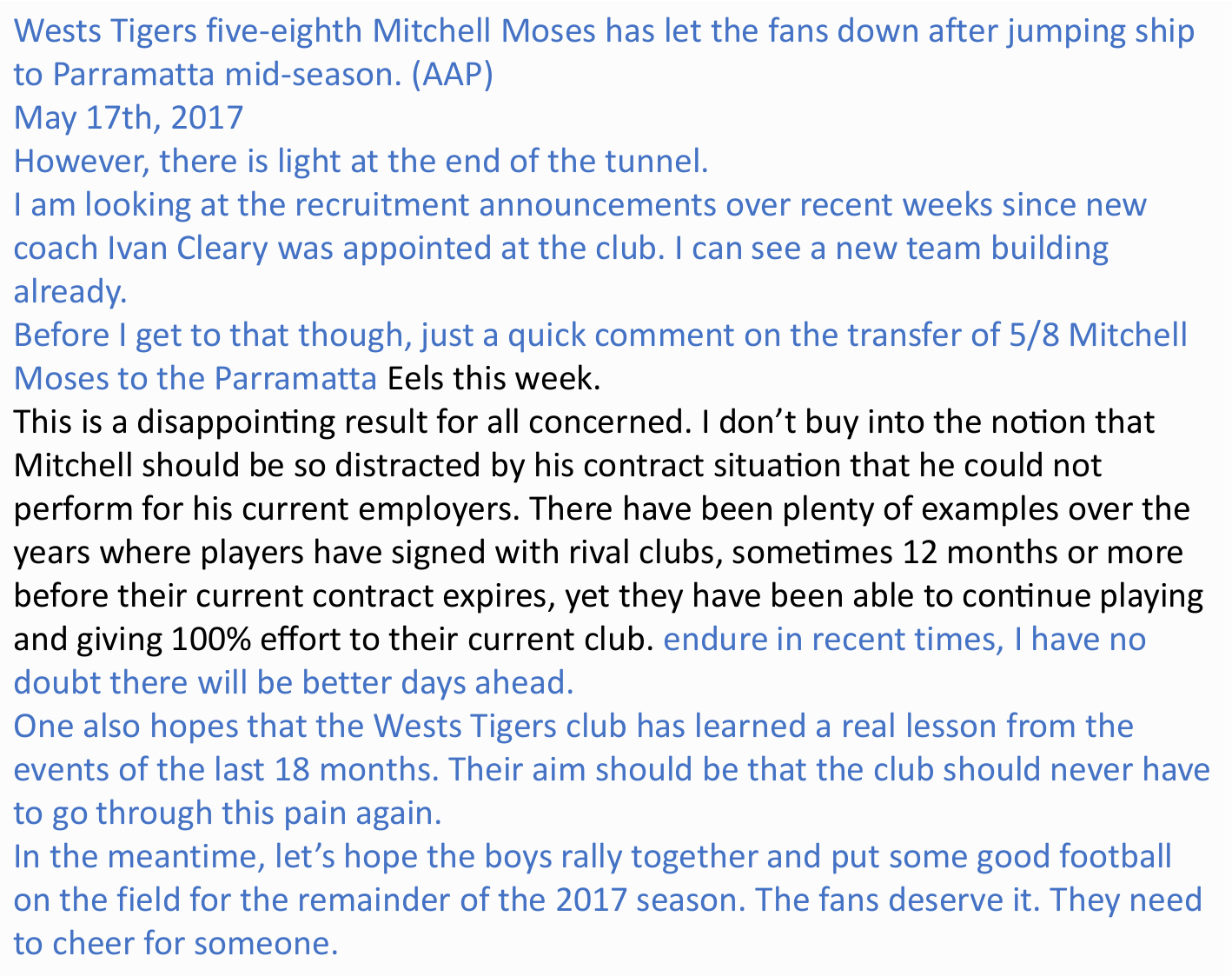}
\caption{Qualitative examples of the text infilling task. Text in {\color{blue}blue} represents the provided prefix and suffix, while text in {\color{black}black} denotes the model's generated results.}
\vspace{-8pt}
\label{fig:infilling_examples_v2}
\end{figure*}

\end{document}